\documentclass[runningheads]{llncs}

\usepackage{eccv}

\usepackage{eccvabbrv}
\usepackage{comment}

\usepackage{graphicx}
\usepackage{booktabs}
\usepackage{wrapfig}

\usepackage[accsupp]{axessibility}  % Improves PDF readability for those with disabilities.

\usepackage{hyperref}

\usepackage{orcidlink}
\usepackage{amsmath,amssymb}

\usepackage{notation}
\usepackage{listings}[language=Python]
\usepackage{makecell}
\usepackage{multirow}
\usepackage{color, colortbl}

\begin{document}

% ---------------------------------------------------------------
% TODO REVIEW: Replace with your title
\title{Skews in the Phenomenon Space Hinder Generalization in Text-to-Image Generation} 

% TODO REVIEW: If the paper title is too long for the running head, you can set
% an abbreviated paper title here. If not, comment out.
\titlerunning{Skews in the Phenomenon Space Hinder Generalization in T2I}

% TODO FINAL: Replace with your author list. 
% Include the authors' OCRID for the camera-ready version, if at all possible.
%\author{Anonymous Submission}
%\institute{}
\author{
\begin{center}
    \vspace{-15pt}
    Yingshan Chang$^{1}$ \hspace{2em} 
    Yasi Zhang$^{2}$ \hspace{3em} 
    Zhiyuan Fang$^{3}$\\
    \vspace{2pt}
   Ying Nian Wu$^{2}$ \hspace{3em} 
   Yonatan Bisk$^{1}$ \hspace{3em} 
   Feng Gao$^{3}$ \\
   \vspace{-10pt}
   \end{center}
}

%\author{Yingshan Chang\inst{1}\orcidlink{0000-1111-2222-3333} \and
%Second Author\inst{2,3}\orcidlink{1111-2222-3333-4444} \and
%Third Author\inst{3}\orcidlink{2222--3333-4444-5555}}

% TODO FINAL: Replace with an abbreviated list of authors.
\authorrunning{Y.Chang et al.}
% If there are more than two authors, 'et al.' is used.

% TODO FINAL: Replace with your institution list.
\institute{Carnegie Mellon University \email{\quad yingshac@andrew.cmu.edu}\footnote{\url{https://github.com/zdxdsw/skewed_relations_T2I}}
\and
University of California, Los Angeles 
\and
Amazon\footnote{This work is not related to these authors’ position at Amazon.}
% \url{http://www.springer.com/gp/computer-science/lncs} \and
% ABC Institute, Rupert-Karls-University Heidelberg, Heidelberg, Germany\\
% \email{\{abc,lncs\}@uni-heidelberg.de}
}

\maketitle
\vspace{-18pt}
% \begin{center}
%     \footnotesize{This work is not related to these authors’ position at Amazon.}
% \end{center}

\begin{abstract}
  The literature on text-to-image generation is plagued by issues of faithfully composing entities with relations. But there lacks a formal understanding of how entity-relation compositions can be effectively learned. Moreover, the underlying phenomenon space that meaningfully reflects the problem structure is not well-defined, leading to an arms race for larger quantities of data in the hope that generalization emerges out of large-scale pretraining. We hypothesize that the underlying phenomenological coverage has not been proportionally scaled up, leading to a skew of the presented phenomenon which harms generalization. We introduce statistical metrics that quantify both the linguistic and visual skew of a dataset for relational learning, and show that generalization failures of text-to-image generation are a direct result of   incomplete or unbalanced  phenomenological coverage. We first perform experiments in a synthetic domain and demonstrate that systematically controlled metrics are strongly predictive of generalization performance. Then we move to natural images and show that simple distribution perturbations in light of our theories boost generalization without enlarging the absolute data size. This work informs an important direction towards quality-enhancing the data diversity or balance orthogonal to scaling up the absolute size. Our discussions point out important open questions on 1) Evaluation of generated entity-relation compositions, and 2) Better models for reasoning with abstract relations.

\end{abstract}

\begin{figure}[h]
\centering
  \vspace{-9pt}
  \includegraphics[width=11cm]{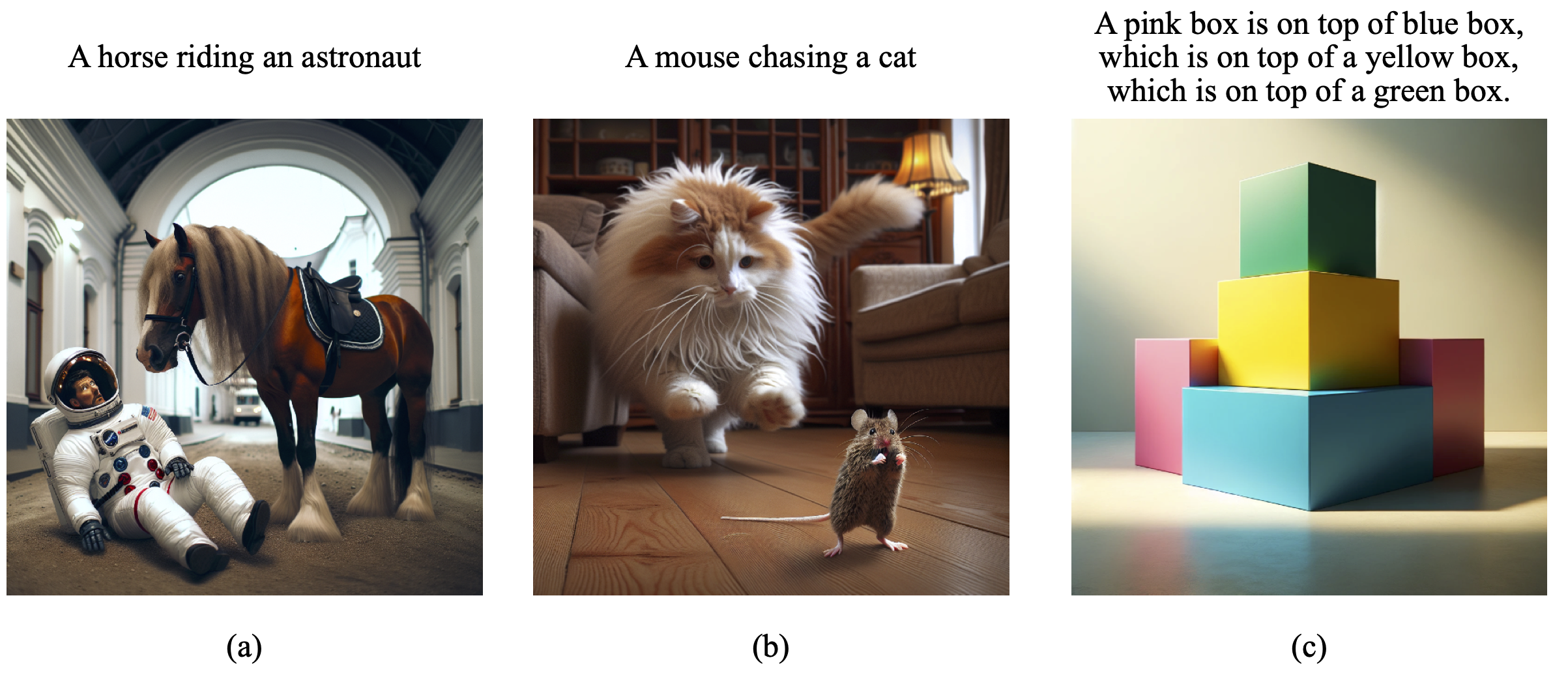}%
  \vspace{-6pt}
  \caption{Example images generated by DALL·E3. In all three cases, entities and relations are common but their compositions are uncommon. DALL·E3 tends to (a) compose entities unnaturally, (b) get trapped by the canonical relation, or (c) disregard the requested ordering. These errors are recurring across multiple trials, suggesting that DALL·E3 does not grasp the abstract notion of relations.}
  \vspace{-18pt}
  \label{figure:dalle3_examples}
\end{figure}

\vspace{-18pt}
\section{Introduction}
\label{sec:intro}

\vspace{-6pt}
A visual scene is compositional in nature \cite{compositionaldiversity}. Atomic concepts, such as entity and texture, are composed via relations \cite{VG}. Relations represent abstract functions that are not visually presented, but modulate the visual realization of concepts. For example, consider a scene “\textit{a cat is chasing a mouse}”. It consists of atomic concepts: cat and mouse. ``Chasing'' defines a relation that is visually realized as certain postures and orientations of the cat and the mouse. Relations take concepts to fill their roles as functions take variables to fill their arguments. This process is known as role-filler binding \cite{smolensky1990, analogy_relational_reasoning, binding_compositional_connectionism, rules_and_variables}, where fillers are concrete values and roles are abstract positions. %Roles have also been referred to as “\textit{scripts}”, “\textit{schema}”, or “\textit{frames}” in studies dating back to the last century \cite{minsky_frame, memory_schema, rules_and_variables}, which is crucial in the context of reasoning with systematic and abstract knowledge. 
%Role-filler binding involves the ability to systematically apply arbitrary fillers to existing roles, leading to novel compositions.  

%Due to abstractness, relations can only be demonstrated and tested with concrete concepts. Therefore, how to truly learn and evaluate relations remains to be an unsolved question. Taking the role-filler binding perspective, truly learning the relation requires learning the bindings independent of fillers they bind. Thus, binding to unseen fillers provides a way to test the success of relational learning. 

\begin{figure}[h]
\centering
  \vspace{-15pt}
  \includegraphics[width=12cm]{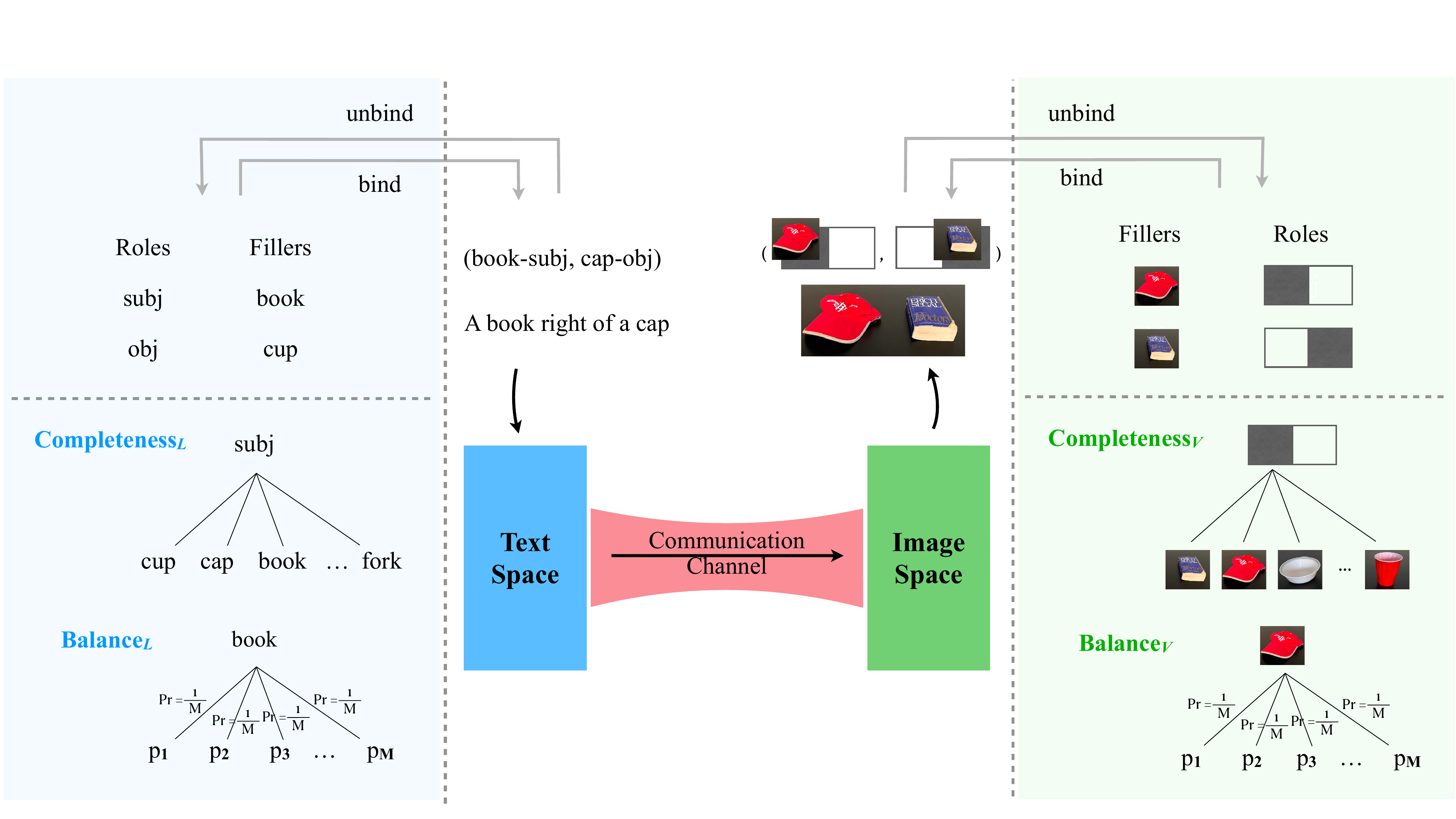}%
  \caption{Conceptual Framework. Text-to-Image generation consists of three important distinct components: A text encoder, a visual decoder, and a mechanism to communicate between these two spaces. Generation of images with consistent spatial relations requires that 1) the text encoder distinctly encodes linguistic roles, 2) the image generator distinguishes spatial roles in the output space, and 3) learning the correct translation from linguistic roles to visual roles. Suppose pre-training or architectural expressivity can fulfill the first two requirements, the remaining core task is to learn an effective communication channel -- often instantiated as cross-attention in diffusion models. To this end, we propose statistical metrics to formally quantify how the training data distribution received by the communication channel affects generalization.}
  \vspace{-6pt}
  \label{figure:intro}
\end{figure}

\vspace{-9pt}
Due to abstractness, relations are always bound to concrete concepts in the space of observations, posting the challenge of truly grasping the abstract function of a relation and using it in generalizable ways, i.e. composing familiar relations with novel concepts. Recently, pre-trained text-to-image models \cite{dalle3, muse, imagen} unleash the power of image synthesis with unprecedented fidelity and controllability. However, as shown by Figure~\ref{figure:dalle3_examples}, a pre-trained model cannot generate images faithful to the relational constraints upon seeing uncommon entity-relation compositions. This implies that, pre-trained text-to-image models do not represent role-filler bindings independently of the fillers, leading to an important question of \textbf{what hinders the learning of generalizable relations}. %\textbf{when and why text-to-image models fail to learn generalizable relations}.

This work investigates this question from the data distribution angle. We conjecture that although pre-training ensures massive data quantity, it does not accomplish a proportionally large coverage of unique phenomena. Figure \ref{figure:intro} shows our conceptual framework for text-to-image generation, consisting of three distinct components: A text encoder, a visual decoder and a mechanism to communicate between these two spaces. We formalize the underlying structure of the data as role-filler bindings which nicely capture the compositional connections between data points. We assume that architectural expressivity and pretraining already enable both the text encoder and the visual decoder to distinctly represent fillers and roles in their corresponding spaces. Based on this assumption, the communication channel becomes the key to task success. We believe the choice of supervision data crucially affects the behavior of the communication channel.

To this end, we introduce two metrics that quantify the skew of the underlying structure supported by a dataset. These two metrics take into account linguistic notion of roles and visual notion of roles respectively. Our hypothesis is that generalization failure of text-to-image model is a direct result of phenomenological incompleteness or imbalance under our metric. Our experiments in both synthetic images and natural images demonstrate the strong predictive power of our metrics on generalization performance. We also show that findings from pixel diffusion models carry over to latent diffusion models.
%To validate this hypothesis, we first present experiments in a synthetic domain where metrics can be systematically controlled. The synthetic experiments demonstrate the strong predictive power of our metrics on generalization performance. In light of our theoretical analysis, we perturb natural data towards a stronger or weaker generalization potential, and observe corresponding shifts in performance in accordance with the theoretical prediction. 
%This work mainly focuses on spatial relations, but extending our framework to the relational realization of verbs more generally is an important future direction. 

% The remaining sections are arranged as follows: Section 2 discusses related work. Section 3 mathematically defines the proposed metrics for linguistic and visual skew. Section 4 discusses experimental results in synthetic domain. Section 5 presents experiments on naturalistic data, with implications on how to enhance a dataset along axes orthogonal to scale. We conclude with discussions by pointing out issues with text-to-image evaluation methods which might hide the weak relational learning ability, and by calling for new architectures that more effectively learn role-filler bindings even from skewed data sources. 

\section{Related Work}
\label{sec:literature}
\textbf{Text Conditioned Image synthesis}
Diffusion models initiate the tide of synthesizing photorealistic images in the wild. They benefit from training stability and do not exhibit mode collapse that GAN models suffer from. \cite{ControlNet} feeds text prompts to the diffusion model to make the generation process controllable. Inspired by ControlNet, a myriad of works \cite{freecontrol, plugandplay, parti, GLIGEN, imagen, dalle2} explore the integration of text encoders and image generators, such that image synthesis, editing and style-translation can be customized by users via text. Unlike diffusion models, Transformer-based image synthesis models are naturally better at working in coordination with text, due to the shared tokenization process \cite{muse, LQAE}. Transformer-based approaches perform on par with diffusion on fidelity, and are believed to have greater potential for resolving long-range dependency and relational reasoning \cite{diffusiontransformer}, thanks to their patch-based representations and attention blocks. However, Transformers suffer from a discrete latent space and slow inference speed.
%discrete nature seems to be at odds with continuity of the visual space. Also, compared with diffusion, Transformers’ much slower inference speed also impedes the deployment in real applications. 
The latest work \cite{diffusiontransformer} integrates a Transformer architecture and diffusion objectives, aiming at the best of both worlds. 

Despite high fidelity scores, a recurring problem is the difficulty to generate objects in unfamiliar relations \cite{trainingpriors}. Although those unfamiliar relations rarely occur in a natural collection of images, they are not physically implausible, and humans have no trouble producing a corresponding scene. This has drawn attention to evaluating generative models and characterizing such failures. Works along this vein suggest that generative models typically fail at multiple objects, with multiple attributes or relations \cite{VISOR, T2I-compbench}, where generalization to novel combinations of familiar components is needed the most.

\vspace{5pt}
\noindent\textbf{Compositional generalization in image synthesis}
Compositional generalization is a specific form of generalization where individually known components are utilized to generate novel combinations. This remarkable learning ability has been widely studied in the vision-and-language understanding domain \cite{emergentcompositionalsolvingmath, fertility, pottscompositionality, SCAN, gSCAN, ReaSCAN, clevr}. The text-to-image literature has recently seen efforts towards constructing images compositionally \cite{scenecomposer, fastcomposer, pairdiffusion}. Closest to ours are two previous works that characterize properties of the underlying phenomenon space not trivially revealed by the pixel space. \cite{cube} has investigated shape, color and size as domains of atomic components, which can be combined to form tuples, e.g. (big, red, triangle). They mainly argue that generalization occurs under two conditions: 1) 
%A novel tuple differs from some training tuple by one element
small structural distance between training and testing instances, and 2) effectively learning the disentanglement of attributes (i.e. a change in the size input will not affect the color output). \cite{firstprinciple} first assumes compositional data are formed by combining individually complex components with simple aggregation functions. Then they defined \textit{compositional support} and \textit{sufficient support} over a set of components, which are sufficient conditions for a learning system that compositionally generalizes.

%\noindent\textbf{Architectures for Learning Generalizable Relations}
%Generalization imposes requirements not only on data conditions, but also on architectures so as to have the capacity in theory. The ability to process relations between inputs \cite{simplerelnet, abstractors} is believed to be a crucial condition. \cite{similaritymatrix} proposed a relational-learning network that explicitly processes the affinity matrix among input entities. \cite{vectordifference} modeled vector differences between input features, resulting in reltional representations amenable to recombination of parts. A role-filler binding mechanism is another requirement for nontrivial generalization, especially for learning abstract roles (e.g. positions, syntax) \cite{smolensky1990, tensorproductdecomposition, tensorproductbasic}. \cite{TP-transformer} learned role and filler representations independently and implements the binding process as tensor products. \cite{ESBN} imposed independence between fillers and roles by forcing the relational reasoning stream to be separated from the sensory information stream. 

%All architectures reviewed here fall under the general category of neuro-symbolic models. \cite{neurocompositional} argued that neuro-symbolic architecture has the potential to harness the power of both worlds: differentiability of neural methods and generalizability of symbolic methods. 
Motivated by failures of existing methods, we investigate data-related factors that affect the generalization performance. There is a possibility that better architectural design can complement high-quality data to achieve generalization.

\begin{figure}[t]
\centering
    \vspace{-15pt}
  \includegraphics[width=12cm,height=7cm]{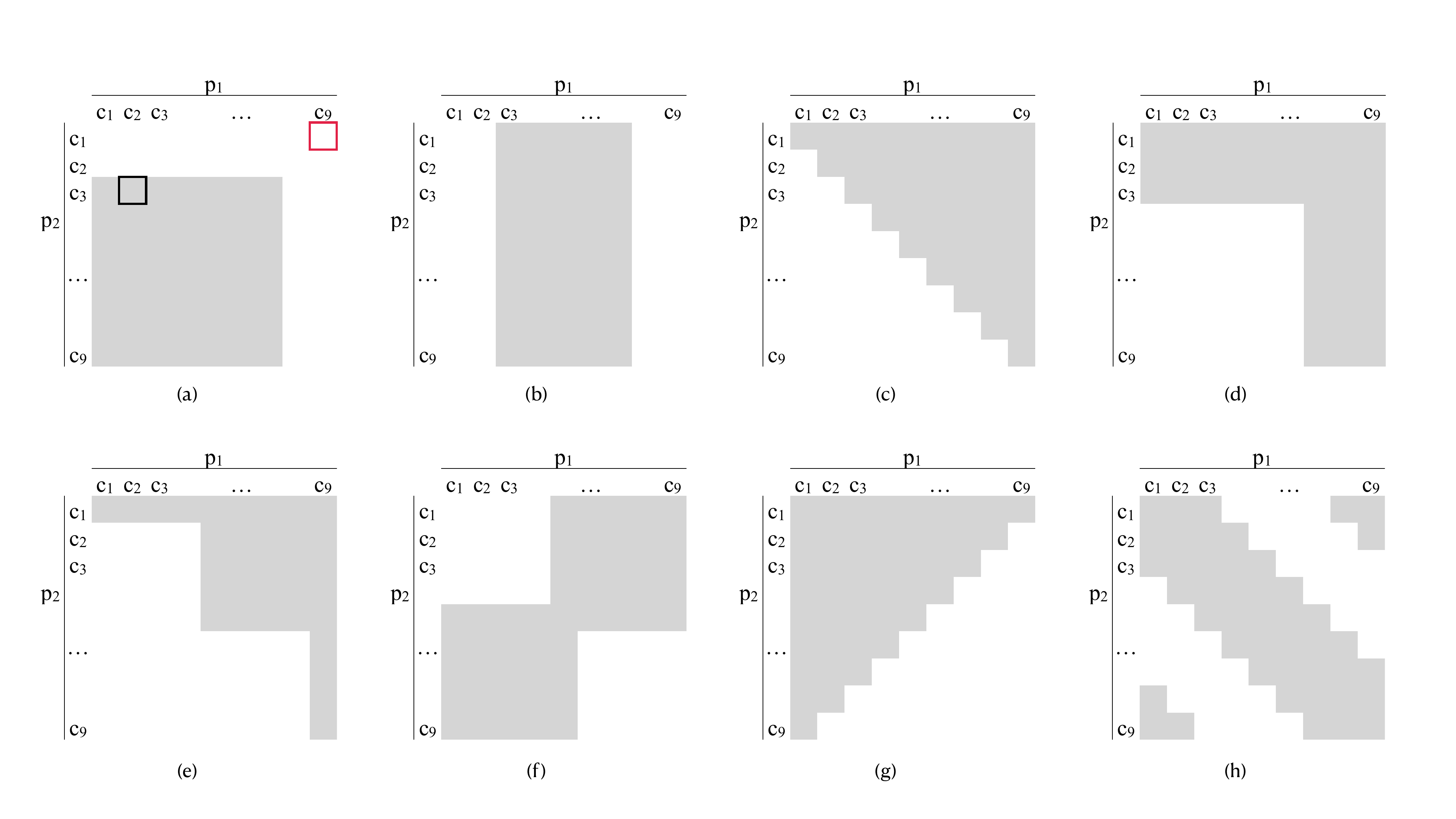}
  \vspace{-10pt}
  \caption{Sketched illustrations of phenomenological coverage with different properties. Shaded areas represent the training set, while blank areas represent the testing set. Columns and rows are organized by the concepts bound to position 1 and position 2 respectively. For example, the black cell in (a) represents the training instance ($c_2/p_1, c_3/p_2$), the red cell in (a) represents the testing instance ($c_9/p_1, c_1/p_2$). (a) Both positions are incomplete (b) Only $p_1$ is incomplete (c) Complete but unbalanced (d) Complete but unbalanced (e) Complete but unbalanced (f) Complete and balanced (g) Complete and balanced (h) Complete and balanced}
  \vspace{-9pt}
  \label{fig:skew_simplified_drawing}
\end{figure}

\section{Formalization}

 {We start by formalizing scene construction as role-filler bindings.  A scene is constructed by binding fillers denoted as $\Fb = (f_1, \ldots, f_K)$ to roles denoted as $\Rb = (r_1, \ldots, r_K)$. Roles and fillers are paired up by their indices. Hence, each scene representation involves the same number of roles and fillers, i.e.,   $|\Fb| = |\Rb| = K$. Using $\psi$ to denote the \textit{binding} operation, a scene can be formalized into: $\psi(\Fb, \Rb) = (f_1/r_1, ..., f_K/r_K)$ and we call $(f_k, r_k)$  a role-filler pair.  } 

 {We would \textit{unbind} a filler from $\psi(\Fb, \Rb)$ via the unbinding operation $\psi^{-1}$: $\psi^{-1}\Big(\psi(\Fb, \Rb), f_k\Big) = r_k$. The unbinding operation describes the process of extracting the role from a binding that a given filler has been bound to, which corresponds to the decomposition of a compositional structure. Assume that $\psi^{-1}$ returns $null$ if the input filler $f_k$ does not exist in the scene.}

 {
Fillers are atomic concepts that can be selected from a set of concepts $\cC = \{c_1, ..., c_N\}$ while roles can take values from a set of candidate positions $\cP = \{p_1, \ldots, p_M\}$. Typically, roles can have intrinsic meanings independent of the meanings of fillers \cite{smolensky1990}. For instance, the meaning of ``upper position" is determined by the y-axis in a 2D image coordinate system, which exists independently of the specific pixel values that fulfill this role in each image. The meanings of roles can be either learned from the task structure or manually designed. In the text-to-image case, the task structure naturally invites two ways to define roles, corresponding to the linguistic and the visual space, respectively. From the linguistic perspective, we consider grammatical positions, e.g. $\cP_L = \{ subject,  object\}$. From the visual perspective, we consider spatial positions, e.g.  $\cP_V = \{ top,  bottom\}$.}

 {
Under our definitions, each image is a scene. Therefore, an image dataset can be essentially abstracted as a collection of bindings: $\cD = \{\psi(\Fb^i, \Rb^i)\}_{i=1, ..., |\cD|}$, where $\Fb^i$ and $\Rb^i$ denotes the fillers and roles in the $i$-th image.  } 
Let $\cU = \cC \times \cP$ be the universe of all possible bindings. The vanilla notion of coverage supported by a dataset is the proportion of $\cU$ that has non-zero supporting examples in the dataset: $\textbf{Coverage}(\cD) = {|Deduplicate(\cD)|}/{|\cU|}$, where the $Deduplicate$ removes examples with equivalent role-filler representations.
We argue that this metric overlooks how elements in $\cU$ are structurally connected. For instance, each element in $\cU$ shares a common role or filler with other elements. Without taking this structural property into account, truly meaningful coverage of diverse and unique phenomena might be conflated by the seemingly diverse surface forms.  

{Motivated by this consideration, our proposed metrics aim to measure whether a dataset has support for every concept occurring in every position, as well as the probability distribution of the positions that each concept has been bound to. 
  Next, we formally describe completeness and balance metrics by conveniently leveraging the notations of binding and unbinding operators. }

\subsection{Completeness} 

 {Completeness requires that every relation has been bound with every concept across the entire dataset.  In this sense, we define:}
\begin{equation}
    \textbf{Completeness}(p_m, \cD) = \frac{\Big|\Big\{c_n ~ \Big|p_m \in  \psi^{-1} (\cD, c_n), c_n \in \cC   \Big\}\Big|}{|\cC|},
\end{equation}
 where  the operator extracting a concept from a dataset is defined as: 
\begin{equation}
    \psi^{-1} (\cD, c_n)  = \bigcup_{i = 1}^{|\cD|} \Big \{  \psi^{-1}\left (\psi(\Fb^i, \Rb^i), c_n \right) \Big\}.
\end{equation}
A fully complete dataset should have completeness scores of 1 for all relations. 
We take the expected value to obtain an aggregated score over the entire dataset:

\vspace{-6pt}
\begin{equation}
 \begin{split}
    \textbf{Completeness}(  \cD) &= \E \Big[ \textbf{Completeness}(p_m, \cD)\Big] \\
    &= \sum_{p_m \in \cP}\PP(p_m) \textbf{Completenss}(p_m, \cD).
\end{split}
\end{equation}

\subsection{Balance}
 {Balance requires that every concept is bound with each position with equal probability. Concretely, we first calculate the entropy of all positions that a given concept $c_n$ was bound to within dataset $\cD$:  }
\begin{equation}
    \textbf{Balance}(c_n, \cD) = Entropy    \Big [   \psi^{-1}\Big(\psi(\Fb^i, \Rb^i), c_n\Big),  {i=1, ..., |\cD|} \Big ],
\end{equation}
where the function $Entropy$ computes the entropy of the distribution of the positions. Note that we constrain the computation to practical positions, i.e., excluding $null$ during the calculation  process. 
 {Then we aggregate by taking the expected value to obtain the balance over the entire dataset:}
 {
\begin{equation}
       \textbf{Balance}(\cD)  = \E \Big[ \textbf{Balance}(c_n, \cD)\Big] = \sum_{c_n \in \cC}\PP(c_n) \textbf{Balance}(c_n, \cD).
\end{equation}}

This metric is upper bounded by $\log(M)$, corresponding to the entropy of a uniform distribution over  $\cP$. The lower the metric, the higher the skew of a dataset.  We normalize by $\log(M)$ to obtain a value within $[0, 1]$. 

So far we have defined completeness and balance for arbitrary sets of fillers and roles. Hereinafter, we will focus on binary relations, i.e. $|\Fb| = |\Rb| = 2$. Subscripts $_L$ and $_V$ denote the linguistic and visual perspectives under which a metric is computed. Practically, metrics are estimated by empirical counts. Next, we conduct experiments to demonstrate that generalization is hindered by incompleteness or imbalance under either perspective.

\begin{figure}[t]
    \centering
    \includegraphics[width=0.8\textwidth]{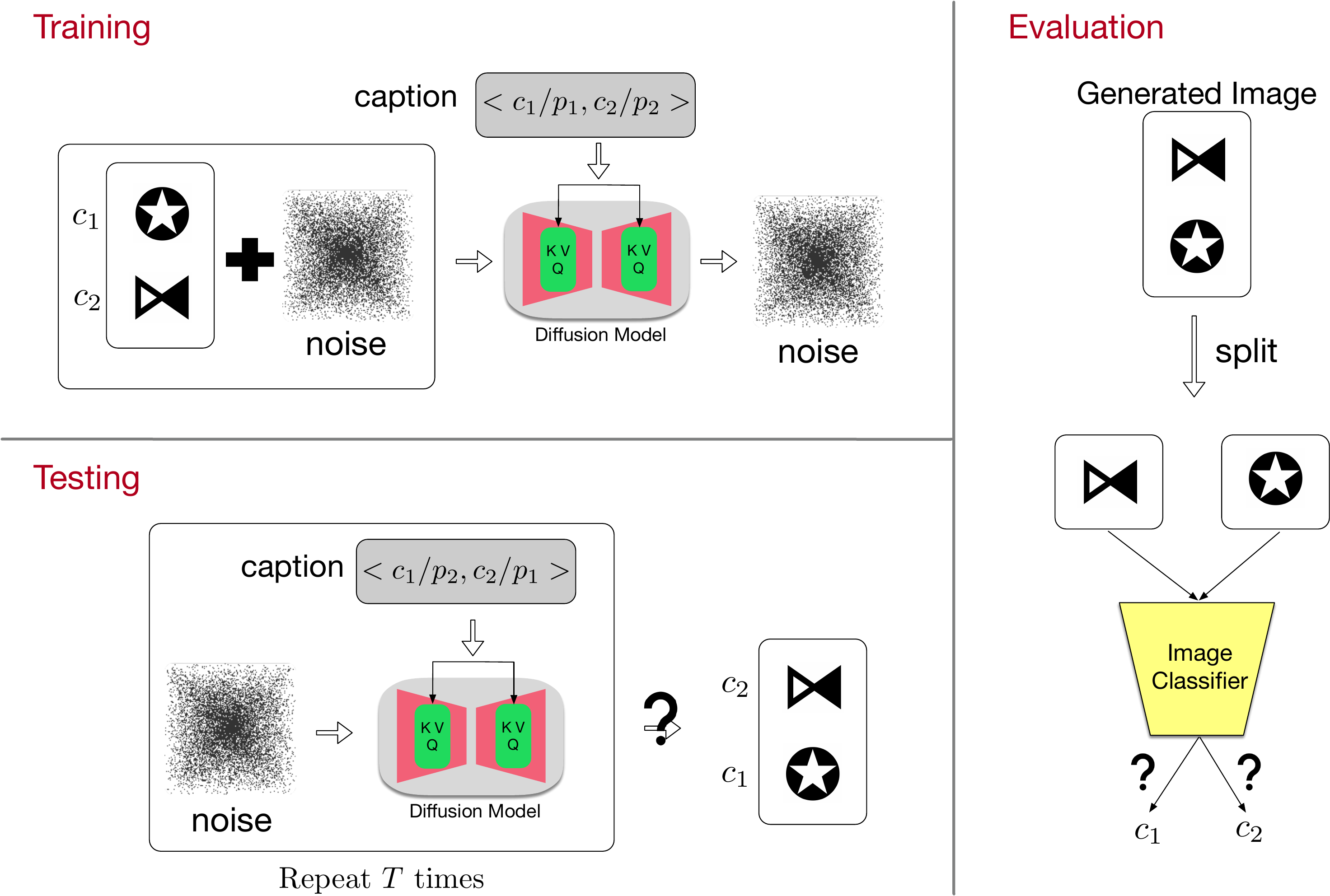}
    \vspace{-3pt}
    \caption{Training, testing and evaluation pipeline. We train diffusion models to generate images of two concepts ($c_1$, $c_2$) with a specified spatial relation. Then the model is tested on unseen concept pairs to see whether the learned relations are generalizable.}
    \vspace{-9pt}
    \label{fig:experiments}
\end{figure}

\begin{figure}[t]
    \centering
    \includegraphics[width= 0.9\textwidth]{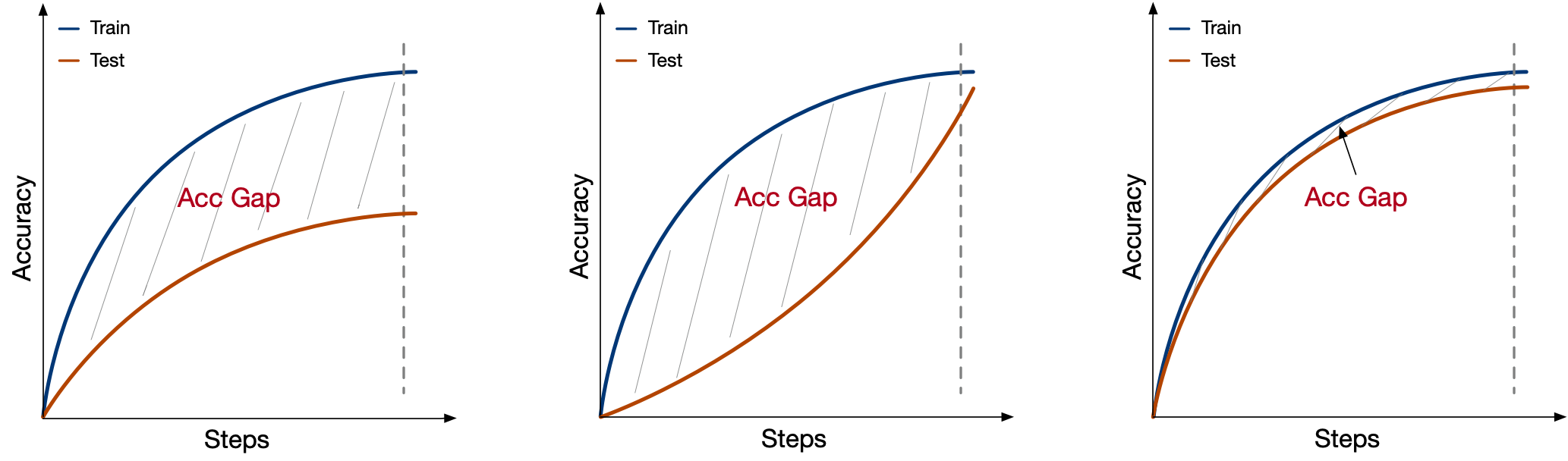}
    \caption{Three types of learning dynamics are observed in our experiments. In the worst scenario (left), the testing accuracy plateaus and never converges. In the best scenario (right), the testing accuracy closely tracks the training accuracy until both converge to perfect. In the middle scenario (center), the testing accuracy climbs slower than the training accuracy, but is still able to converge to perfect at a delayed point after the training accuracy has already converged. In order to distinguish between the middle and best scenarios, we additionally report the accumulative gap between training and testing accuracy curves, which captures the timeliness of generalization.}
    \vspace{-12pt}
    \label{fig:train_test_gap}
\end{figure}

\section{Experiments on Synthetic Images}
\label{sec:experiments_on_symthetic_images}

\subsection{Setup}

We use a set of unicode icons as concepts, and assign common nouns to them as their names. We vary the number of concepts, $N$, within $\{30, 40, 50, 60, 70, 80, 90\}$. We consider two symmetric binary relations: "on top of", "at the bottom of". Images are constructed by drawing each icon on a 32x32 canvas, then stacking two icons vertically, resulting in 32x64 resolution. Captions are created from the template: "a(an) <icon\_name> is <relation> a(an) <icon\_name>".

Training sets ($\cD$) are sampled from $\cU$ with systematic control for the four properties: \textbf{Completeness}$_L$, \textbf{Completeness}$_V$, \textbf{Balance}$_L$, \textbf{Balance}$_V$. To avoid confounders, we ensure the linguistic metrics are perfect when studying the effect of the visual metrics, and vice versa. Figure \ref{fig:skew_simplified_drawing} illustrates training distribution with varied properties. We take the complementary set, $\cU\setminus\cD$ as the testing set. Testing instances are unseen in terms of the tuple representation $(f_1, r_1, f_2, r_2)$. Yet we show that perfect testing accuracy is possible when the training set properly supports the phenomena space, i.e. being complete and balanced. 

Our training, testing and evaluation pipeline is depicted in Figure \ref{fig:experiments}. Following the architecture of \cite{ddpm}, we train 350M UNet models on text-conditioned pixel-space diffusion, with T5-small \cite{T5} as the text encoder. The size of UNet is determined such that it is minimally above the threshold at which the model is capable of fully fitting the training set. This would avoid issues such as a lack of expressivity or under-training of an overparameterized model.\footnote{We adopt an lr of 1e-4 and batch size of 16. Evaluation is performed every 20 epochs. Early stopping is applied when the evaluation has not been better for 100 epochs. The length of training typically falls between 600 epochs ($N=30$) and 200 epochs ($N=90$).  A full list of model and training configs is provided in the appendix.}
For evaluation, we compute accuracy of both icons being generated correctly\footnote{During evaluation, we find that the random state at which each diffusion process begins with has negligible effect ($\pm1\%$) on final performance}. The evaluation process is automated by pattern-matching icons with convolutional kernels. 

Alongside the final testing accuracy, we report the accumulative difference between training and testing accuracy curves. This is because we have observed a third type of learning dynamics lying in between a generalization success and generalization failure (Figure \ref{fig:train_test_gap}), where the testing accuracy climbs slower than the training accuracy, but is still able to converge to perfect at a delayed point
%after the training accuracy has already converged. 
Only reporting on final testing accuracy would mask the qualitative difference that captures a notion of how timely generalization occurs. 

\subsection{Results}
\label{sec:synthetic_results}

\begin{figure}[t]
    \centering
    \vspace{-3pt}
    \includegraphics[width=0.75\textwidth]{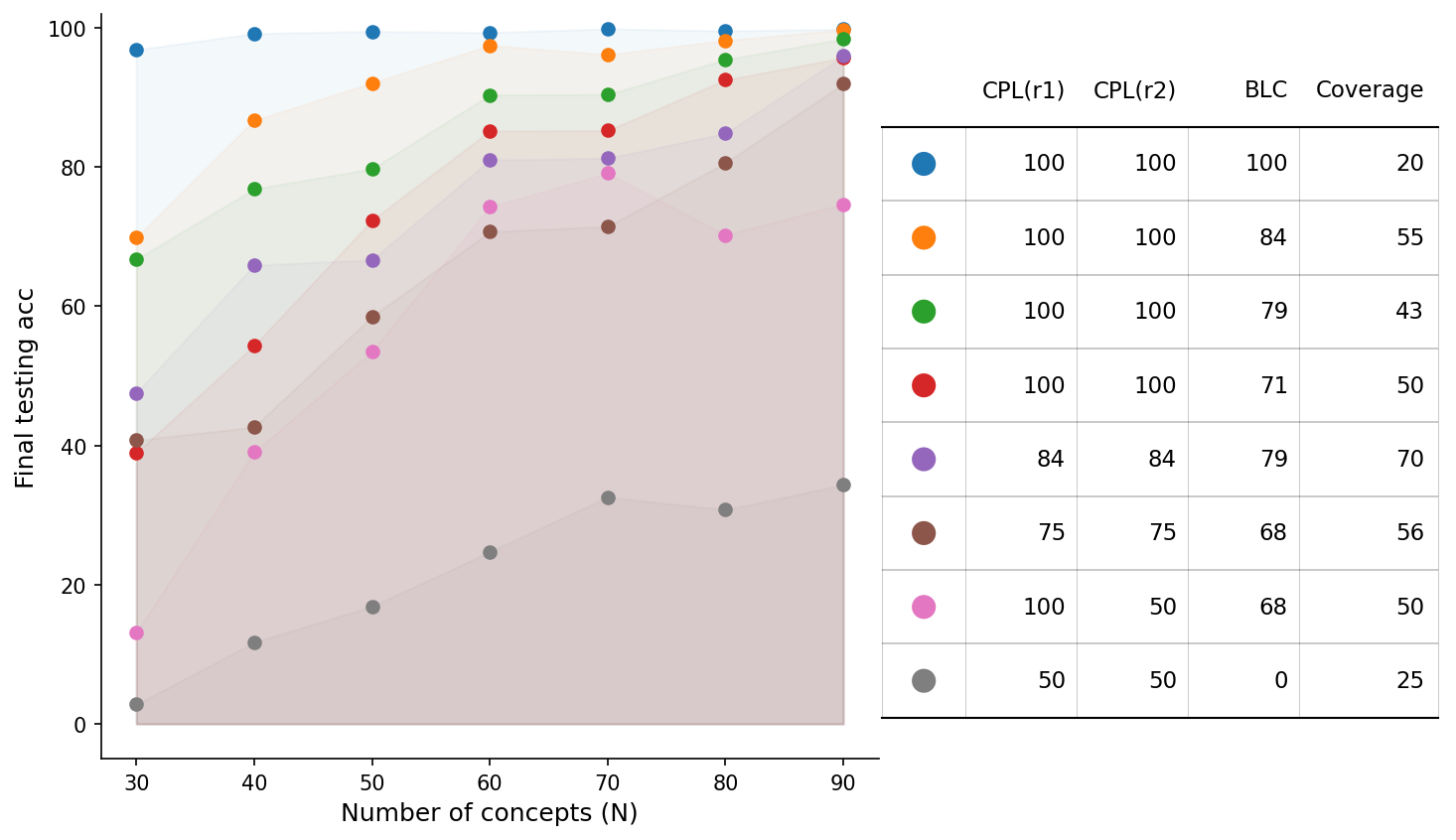}
    \vspace{-3pt}
    \caption{Under the definition of roles ($r_1$, $r_2$) from \textbf{visual} perspective (``top", ``bottom"), the plot (left) shows final testing accuracy against distributional properties of the training set. The legend and the corresponding metrics are summarized in the table (right). The results suggest that both \textbf{Completeness$_V$} (CPL) and \textbf{Balance$_V$} (BLC) are positively correlated with testing (generalization) performance. By contrast, a vanilla notion of data coverage is badly correlated with performance.}
    \vspace{-15pt}
    \label{fig:image_skew_progression_and_table}
\end{figure}

\noindent\textbf{Visual incompleteness significantly impedes generalization.}
The last four experiments in Figure \ref{fig:image_skew_progression_and_table} (indexed by purple, brown, pink and grey) have low \textbf{Completeness}$_V$. Their testing performance never reached 100\%, even plateauing below 50\% for smaller concept sets. 

\vspace{3pt}
\noindent\textbf{Visual imbalance harms generalization when $N$ is small.} The first four experiments in Figure \ref{fig:image_skew_progression_and_table} (indexed by blue, orange, green and red) show the progression of increasing \textbf{Balance}$_V$ while keeping full \textbf{Completeness}$_V$. As \textbf{Balance}$_V$ grows, the testing accuracy consistently improves for all $N$. Increasing $N$ can provide a remedy for a dataset with complete but imbalanced support. In contrast, larger $N$ does not bring much help the support is incomplete.

\begin{figure}
\centering
\begin{minipage}{.5\textwidth}
  \centering
  \includegraphics[width=.95\linewidth]{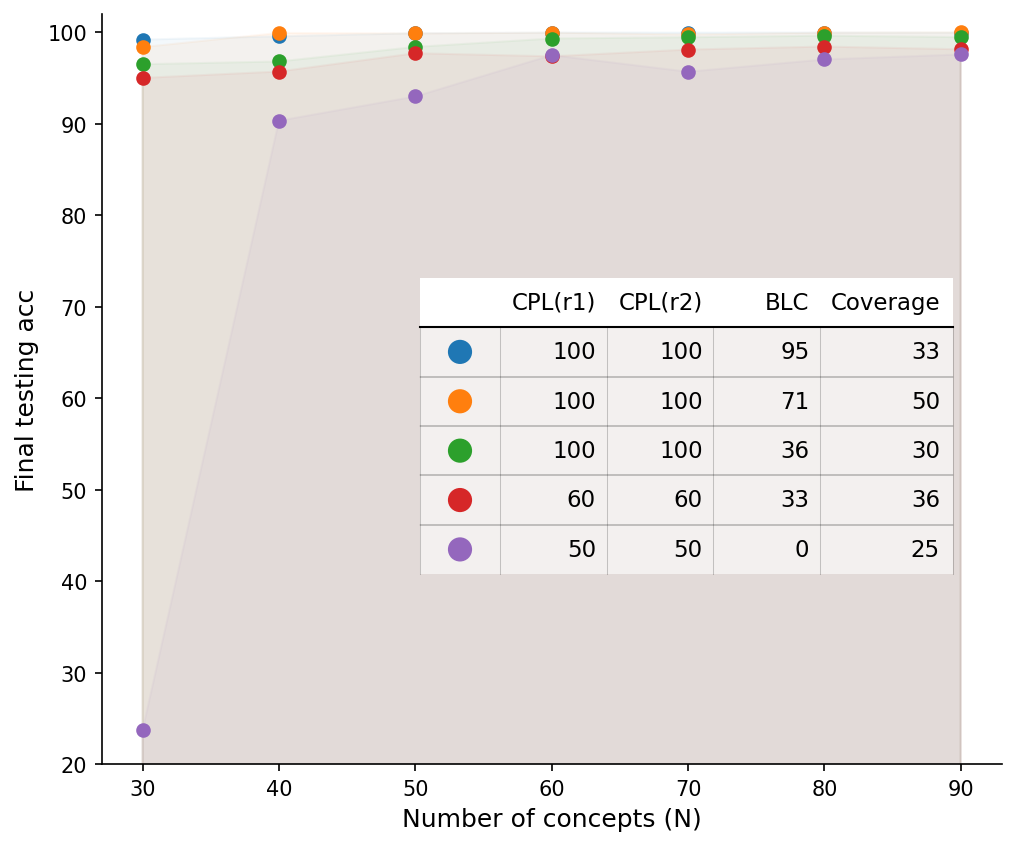}
  %\label{fig:linguistic_skew_progression_and_table}
\end{minipage}%
\begin{minipage}{.5\textwidth}
  \centering
  \includegraphics[width=.95\linewidth]{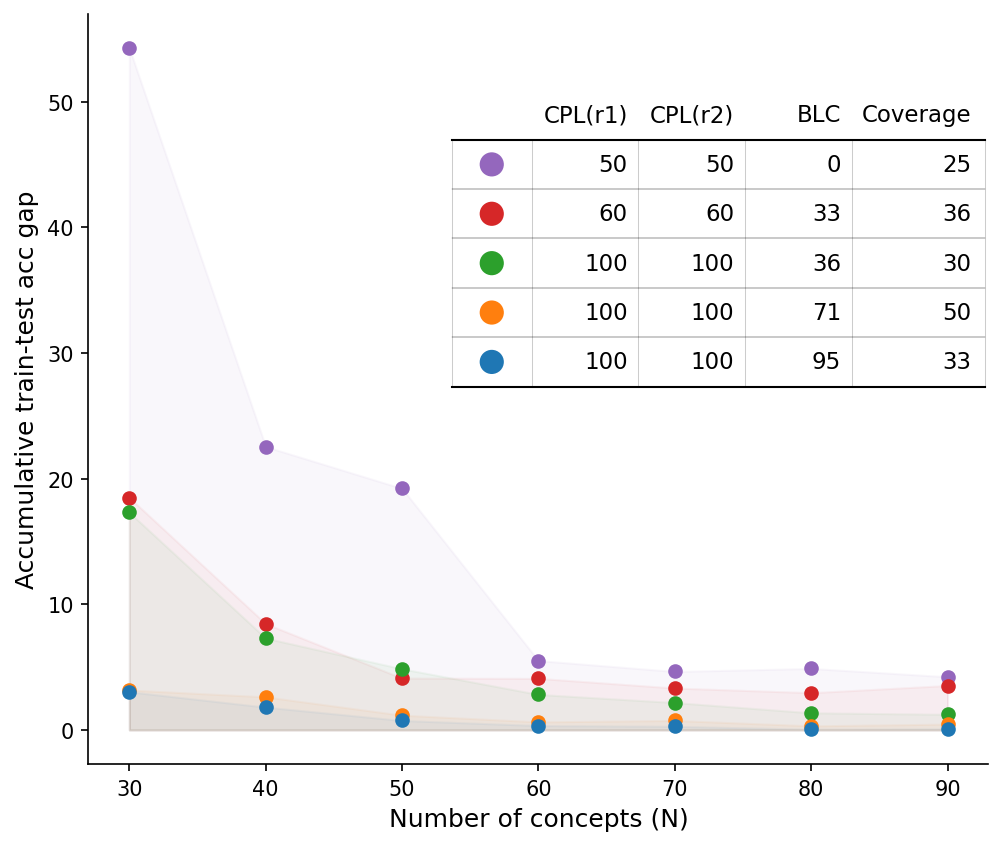}
  
  %\label{fig:linguistic_skew_gap_and_table}
\end{minipage}
\caption{Under the definition of roles ($r_1$, $r_2$), \textbf{linguistic}ally (``subj", ``obj"), the plot shows final testing accuracy (left) and train-test accuracy gaps(right) against distributional properties of the training set. Legend and metrics are summarized in the table. The left plot suggests that linguistic incompleteness and imbalance harm generalization for small concept classes, while having less impact on the final testing accuracy for large concept classes. The right plot suggests that linguistic incompleteness and imbalance harm the timeliness of generalization, as indicated by larger train-test accuracy gaps.}
\vspace{-6pt}
\label{fig:linguistic_skew_side_by_side}
\end{figure}

\vspace{3pt}
\noindent\textbf{Linguistic incompleteness or imbalance harms generalization to a lesser degree, but they delay generalization.} 
Figure \ref{fig:linguistic_skew_side_by_side} (left) shows that all cases achieve perfect or near-perfect testing accuracy, unless for the very small concept classes. This suggests that linguistic incompleteness and imbalance do not severely hinder whether the model is able to generalize ultimately. However, they do bring a negative effect by delaying the onset of generalization. This delay effect is revealed in Figure \ref{fig:linguistic_skew_side_by_side} (right). The takeaway is that, although the final testing accuracy is comparable, lack of $\textbf{Completeness}_L$ or $\textbf{Balance}_L$ causes the testing acc to largely lag behind training acc, which takes a longer time to catch up. Similarly, we plot the generalization gap for the set of visual skew experiments in the appendix, observing the same trend. However, since both failing to generalize or having a prolonged duration before generalizing can lead to a large gap, this result has to be taken with a grain of salt.

\begin{comment}

\begin{figure}[t]
    \centering
    \includegraphics[width=0.85\textwidth]{figures/image_skew_gap_and_table.png}
    \caption{Under the definition of roles ($r_1$, $r_2$) from \textbf{visual} perspective (``top", ``bottom"), the plot shows final testing accuracy against distributional properties of the training set. Legend and corresponding metrics are summarized in the table. The results suggest that linguistic incompleteness and imbalance harm the timeliness of generalization, as indicated by larger train-test accuracy gaps.}
    \label{fig:image_skew_gap_and_table}
\end{figure}
\end{comment}

\vspace{5pt}
\noindent\textbf{Vanilla notions of coverage are a bad predictor for generalization.} The rightmost columns in Figures \ref{fig:image_skew_progression_and_table} and 
\ref{fig:linguistic_skew_side_by_side} provide the training set's coverage (\%) over the universe $\cU$. It can be seen that this does not correlate with the generalization performance. For example, the green run in Figure \ref{fig:image_skew_progression_and_table} outperformed red, purple, brown and pink runs, while having a much lower coverage than them. Also noteworthy is that we intentionally select low-coverage datasets to demonstrate the fully-complete, fully-balanced case, which achieves perfect generalization for all $N$. This strongly indicates the problem of aiming for a generalizable model by recklessly scaling up coverage. In accordance with research on model bias, we argue that scaling up along the incorrect axes is dangerous because it may aggravate unintended bias, without necessarily benefiting generalization. 

\vspace{5pt}
\noindent\textbf{Increasing $N$ eases generalization in all cases.} The model consistently generalizes better when trained on more concepts, albeit to varying degrees. Figures \ref{fig:image_skew_progression_and_table} suggests that enlarging $N$ is more helpful when the data is within a decent range of \textbf{Completeness}$_V$ and \textbf{Balance}$_V$ (e.g. 70-80).

\section{Experiments on Natural Images}
\label{sec:experiments_on_natural_images}
\subsection{Setup}

We extend our experiments to natural images using the What’sUp benchmark proposed by \cite{whatsup}.\footnote{The What’sUp benchmark provided subsetA and subsetB. We adopt subsetB for our purpose because objects in subsetA have a great disparity in their sizes.} Our hypothesis is that higher Completeness and Balance of the training distribution lead to more generalizable learning outcomes. Each image in the What’sUp benchmark contains two objects, with a caption describing their spatial relations. The spatial relations include two pairs of symmetric binary relations: \textit{left/right of} and \textit{in-front/behind}. Without loss of generality, we study \textit{left/right of}, leaving the exploration of more than two relations for the future. 

From an initial (complete and balanced) set of 308 samples with 15 unique concepts, subsamples are drawn where completeness and balance vary. For fair comparisons, the coverage of all subsamples is relatively the same. We train 470M pixel-space diffusion models on What’sUp image-caption pairs, with 64x32 resolution, 5e-4 learning rate and a batch size of 16. Early stopping is applied similarly to the synthetic setting. The length of training typically falls between 3000 and 6000 epochs. Hyperparameter tuning is described in the appendix.

Since under the natural image setting the same object may occur at different image positions, evaluation could not be performed with predetermined pattern-matching kernels. We finetune ViT-B/16 \cite{ViT} as an automatic evaluation engine to classify the object in the left or right crop of generated images.\footnote{The auto-eval engine can be deemed oracle. On 144 manually checked samples, the ViT's judgments are all correct} A ``blank'' label is added to the classifier in order to indicate when a model fails to generate an object. Importantly, a secondary result of our work is demonstrating shortcomings of the widely used evaluation methods for image synthesis, such as CLIPScore \cite{clipscore}, VQA with VLMs \cite{llmscore, T2I-compbench}, bounding-box evaluation with detectors \cite{VISOR, T2I-compbench}. See Section \ref{sec:discussions} for discussions of when they fall short.

\begin{table}[t]
\parbox{.48\textwidth}{
\centering
  \caption{\footnotesize What’sUp benchmark results varying \textbf{Completeness}$_V$ and \textbf{Balance}$_V$}
\centering	
\small
 \begin{tabular}{@{}cccccc@{}}
\toprule
\multicolumn{4}{c}{\footnotesize Training Set Properties} & \multicolumn{2}{c}{\footnotesize Performance}
\\
 \cmidrule(lr){1-4} \cmidrule(lr){5-6}   
{ \tiny CPL($r_1$)}  &  \tiny CPL($r_2$) & \tiny BLC  &   \tiny Cov.  & \tiny Final Acc$^\uparrow$ &  \tiny  Acc Gap $^\downarrow$ \\
  % $r_1$ &  $r_2$ & \tiny BLC & \tiny Cov. &  $^\uparrow$ &   $^\downarrow$ \\
\midrule
     100 &       50 &   63 &        47 &        18.75 &                              84.55 \\
      80 &       73 &   77 &        50 &        19.52 &                              73.87 \\
      87 &       87 &   75 &        49 &        25.93 &                              68.41 \\
     100 &      100 &   73 &        44 &        10.17 &                              80.49 \\
     100 &      100 &   88 &        49 &        28.50 &                              64.89 \\
     100 &      100 &  100 &        48 &        48.64 &                              33.67 \\
\bottomrule
\end{tabular}
  \label{tab:whatsup_Visual}
}
\hfill
\parbox{.48\textwidth}{
\centering
  \caption{\footnotesize What’sUp benchmark results varying \textbf{Completeness}$_L$ and \textbf{Balance}$_L$}
    \label{tab:whatsup_Linguistic}
    \centering
    \footnotesize
 \begin{tabular}{@{}cccccc@{}}
\toprule
\multicolumn{4}{@{}c@{}}{\footnotesize Training Set Properties} & \multicolumn{2}{c}{\footnotesize Performance}
\\
 \cmidrule(lr){1-4} \cmidrule(lr){5-6}   
 { \tiny CPL($r_1$)}  &  \tiny CPL($r_2$) & \tiny BLC  &   \tiny Cov.  & \tiny Final Acc$^\uparrow$ &  \tiny  Acc Gap $^\downarrow$ \\
  % $r_1$ &  $r_2$ & \tiny BLC & \tiny Cov. &  $^\uparrow$ &   $^\downarrow$ \\
  \midrule
      50 &      100 &   63 &        47 &        57.14 &                              40.72 \\
      80 &       73 &   77 &        50 &        60.00 &                              39.83 \\
      87 &       87 &   75 &        49 &        62.04 &                              38.44 \\
     100 &      100 &   73 &        44 &        62.71 &                              33.90 \\
     100 &      100 &   80 &        37 &        65.04 &                              30.21 \\
     100 &      100 &  88 &        49 &        67.29 &                              32.90 \\
\bottomrule
\end{tabular}
}
\end{table}

\begin{figure}[t]
\centering
  \includegraphics[width=\linewidth]{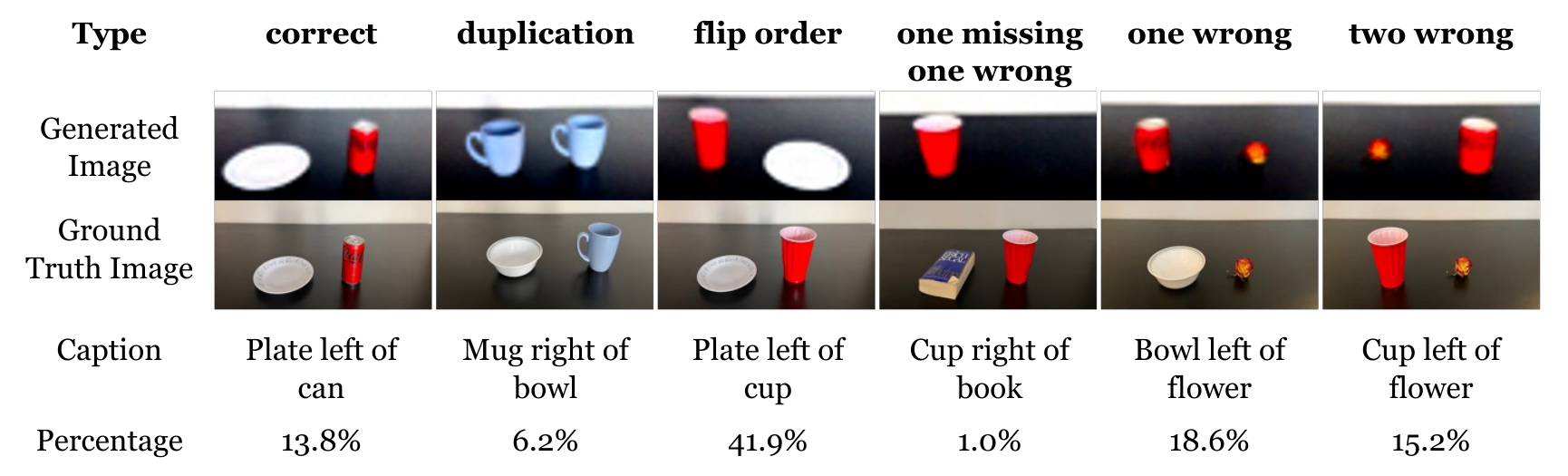}%
  \caption{Qualitative examples of generated images and the corresponding ground truth images, including a categorization of failure types and their frequencies.}
  \vspace{-9pt}
  \label{fig:error_analysis}
\end{figure}

\subsection{Results}

On natural images, it is much harder to generalize, probably because the objects' absolute position and postures can vary even when the relative spatial positions are determined. Another likely cause is the small number of concepts, as suggested by Section \ref{sec:synthetic_results} that generalization tends to plateau at 50\% for a small concept set. Nevertheless, the relative performance across different training set properties still conveys a meaningful message. Table \ref{tab:whatsup_Visual} shows a consistent trend of generalization performance influenced by \textbf{Completeness}$_V$ and \textbf{Balance}$_V$. The final accuracy gets higher and the generalization gap gets lower as the training distribution moves towards being fully complete and balanced. \textbf{Completeness}$_L$ and \textbf{Balance}$_L$ achieve a similar effect, as shown by Table \ref{tab:whatsup_Linguistic}.

As with our previous findings, visual skew imposes a more severe challenge, compared with the same amount of distributional skew on the linguistic side. Our explanation is that, since the network is modeling the pixel space, it is more directly affected by the skew of the observed pixel-space distribution, while skew of the language-space distribution impacts the result rather indirectly.

Upon closer examination of model outputs, the most common error was generating the correct objects with flipped order. This suggests that mapping fillers across domains is easy, but learning to map roles when only observing role-filler bindings is hard. Other errors include generating a blank or duplicating the object. 
%Unlike the flippedOrder error, blank and duplication also occasionally occur when performing inference on the training set, likely as a result of undertraining. 
Figure \ref{fig:error_analysis} visualizes examples of correct and incorrect generations.

\section{Discussions}
\label{sec:discussions}

\vspace{-6pt}
We present a conceptual framework and formal metrics to study the contributing factors of generating images with correct spatial relations. Clearly, our work triggers many open questions that are worth future exploration. Appendix~\ref{sec:limitation_and_future} further discusses limitations of this work and future directions.

\vspace{5pt}
\noindent\textbf{Are spatial relations distinguishable within unimodal spaces?}
We have mainly focused on what enables entity-relation compositions to be successfully conveyed from the text to the image domain. However, this question is meaningful only under the assumption that different roles and fillers are distinctly encoded in unimodal spaces. We have evidence that, perhaps surprisingly, this assumption does not always hold in existing approaches. 

We train probing classifiers to extract the positional role of nouns from text encodings. More details are available in Appendix~\ref{sec:probing}. We find that probes trained with the CLIP text encoder can only overfit the training data, but not generalizing. This indicates the inherent weakness of CLIP text encoder to provide consistent signals of spatial positions. This finding aligns with existing criticism on CLIP essentially being a bag-of-word model \cite{vlm_bow}. By contrast, probing experiments with T5 \cite{T5} encoder and the encoder of a pretrained vision-language model \cite{vlc} succeed with near-perfect generalization. 
This makes T5 and VLMs naturally better candidates for training text-to-image models. 

On the image side, the capacity to represent positions can be theoretically guaranteed by image-patch positional encodings, readily compatible with attention blocks in the diffusion architecture. However, to our surprise, most of the open-source diffusion implementations \cite{hf-diffusers} omit this step. We noticed this issue as our initial experiments failed unexpectedly. The problem was fixed after we modified the architecture to include image positional encodings. We posit that the lack of image positional encodings imposes a representational deficiency\footnote{The zero paddings in convolutional layers can possibly leak positional information, but they need many layers to propagate information from the periphery to the center.} leading to heavy reliance on pixel correlations and unexpected testing behavior. Ablation studies in Appendix~\ref{sec:img_posemb_ablation} support this argument.

In short, we point out the importance of a text encoder that distinguishes positional roles and an image decoder that has the representational power for spatial information. We emphasize that these are not only important preconditions for our main analysis presented in this paper, but should also be crucial considerations in future generative models or models that do spatial reasoning.

\noindent\textbf{Generation in the Latent Space}
There are abundant studies on latent generation methods \cite{stablediffusion, ldm, muse, diffusiontransformer} that are able to achieve higher resolution. However, the progress towards spatially consistent high-resolution synthesis might be hampered when the data coverage of underlying phenomena does not proportionally scale with resolution. To test whether our arguments carry over to latent-space generation, we conduct experiments with a pretrained (frozen) VAE (stabilityai/stable-diffusion-2-1). We increase the input resolution by a factor of four because VAE applies compression. The results (Appendix~\ref{sec:latent_experiments}) match our intuition that a latent space does not affect the validity of our conceptual and formal frameworks. CPL and BLC consistently correlate with generalization performance. Also in accordance with our existing findings, linguistic skew harms final testing performance to a lesser degree, but delays generalization, as evidenced by large acc gaps. Finally, the latent diffusion results again verified that larger coverage cannot compensate for a poor CPL or BLC. 

Two factors may explain the similarity in results between pixel and latent spaces. First, the latent space feature maps have spatial correspondence with the image. Second, the VAE \cite{VAE} does not have a language component, as such the language-to-vision communication channel is captured by diffusion. While better features may be provided by VAE, they lack any cross-modal correspondence. 
%This work takes an initial attempt to formally measure phenomenological coverage and its amount of skew. 
In summary, increasing the phenomenological coverage and increasing resolution are both important. We leave the questions open on extending our formal notions to superresolution models as well as more nuanced relations.

\vspace{5pt}
\noindent\textbf{Text-to-Image Evaluation Methods}
We rely on several heuristics when automating the evaluation of generated images. This was feasible only when 1) objects are center aligned, and 2) the background is clean. Ultimately, we are interested in evaluating relations with cluttered scenes and with greater appearance variability. Besides finetuning a ViT classifier, we have explored existing off-the-shelf models, but found them limited in one way or another. CLIPScore is known to pay less attention to token orders. Indeed, CLIPScore judges correctly in only 37\% of the times in our setting. Evaluating spatial relations by comparing bounding box locations produced by detectors offers a more structured approach, yet it is restricted to the object classes available in the detection pretraining. For example, ``headphones" and ``tape" are classes in the What'sUp benchmark that do not belong to any of the popular detection datasets, rendering most of the detection models inappropriate for our experiments. Open-vocabulary detectors \cite{OWL} circumvents this problem. But practical issues still exist, such as redundancy in the predicted bounding boxes and sensitivity to text prompts. %, since an open-vocabulary detector is essentially matching image regions to text spans. 

Aside from using CLIP or detectors, the literature has also suggested using vision-language foundation models (VLMs) for synthesized images evaluation.
In addition to the apparent drawback of slow inference, it remains questionable whether VLMs comprehend relations in the first place \cite{vlm_bow}. Our attempts at LLaVA- or BLIP2-VQA were unsuccessful for multiple reasons such as inability to recognize spatial relation (Table~\ref{tab:vqa}), and unbalanced precisions across object classes (Table~\ref{table:hard_to_detect_classes}). See Appendix~\ref{sec:autoeval} for more analysis. We hope our investigation calls into question the effectiveness of existing metrics on spatial consistency, which might inadvertently mask the weakness of text-to-image models.

\begin{figure}[h]
\centering
\vspace{-9pt}
  \includegraphics[width=\linewidth]{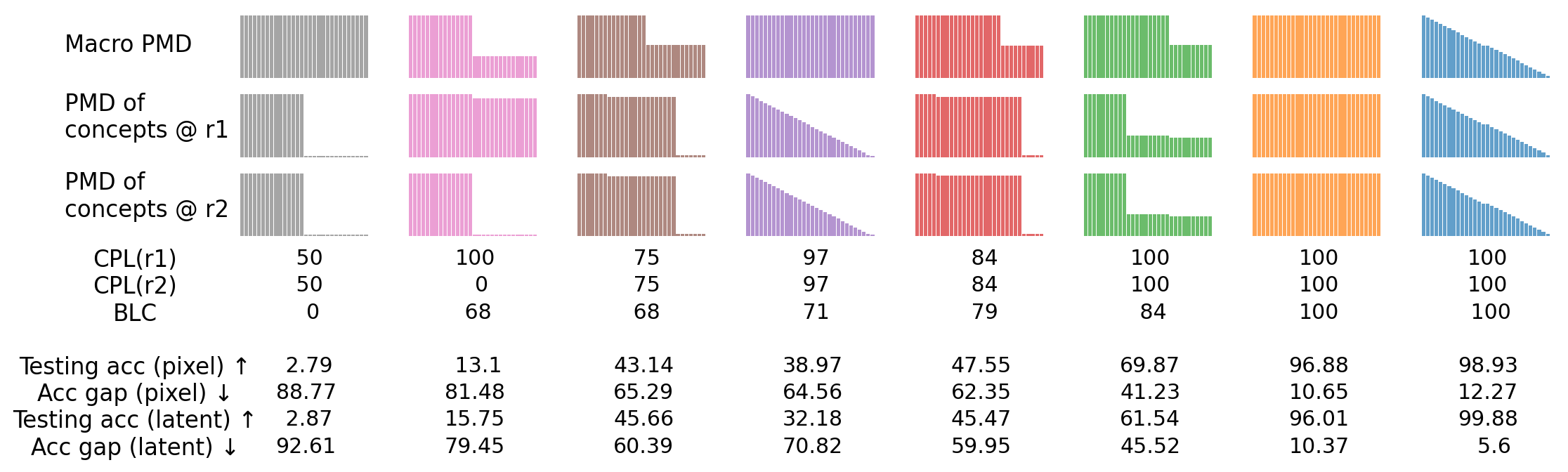}
  \caption{Probability mass distribution (PMD) of various training sets used in Section~\ref{sec:experiments_on_symthetic_images}. Testing performance is correlated more with CPL and BLC, than with the uniformity of PMD. Also note that the PMD of concepts at individual roles may not be uniform although it appears to be uniform on the macro level.}
  \vspace{-18pt}
  \label{fig:pmd}
\end{figure}

\vspace{5pt}
\noindent \textbf{Other data distributional properties} Our metrics are designed around completeness and balance, but they may not capture other distributional properties, such as the Zipfian-Uniform axis. Figure~\ref{fig:pmd} plots the probability mass distribution of datasets we used in Section~\ref{sec:experiments_on_symthetic_images}. We see three shapes: uniform, 2-stages, and ``wedge". This provides new insights: 1) Macro-PMD may hide skew, visible when we plot role-specific-PMD (row 2\&3). E.g. grey and purple instances (col 1\&4) exhibit uniform macro-PMD, but biased role-specific-PMD -- correlated to poor generalization. 2) A non-uniform PMD may not indicate poor generalization. E.g. the blue instance (col 8) has non-uniform PMD, yet achieves great performance -- perfect CPL and BLC scores. Future work may explore other data distributional properties and their correlation with generalization.

\section{Conclusion}

Text-to-Image synthesis, despite recent breakthroughs in fidelity, appearance diversity and texture granularity, still struggles with relations. As the community strives for larger datasets to better cover the natural distribution, there is a lack of study on the axes along which phenomenological coverage can be meaningfully enlarged. This work presents the first effort to formally characterize training coverage, in the context of learning spatial relations. We introduce completeness and balance metrics under both the linguistic and visual perspectives. Our experiments on synthetic and natural data consistently suggest that models trained on more complete and balanced datasets have greater generalization potential. We see this work as a stepping stone towards text-to-image models that can faithfully generate relations in general, including implicit relations entailed by verbs. We hope to inspire research on structured image evaluation, architectures for modeling role-filler bindings and formal frameworks of generalization.

\par\vfill\par

% ---- Bibliography ----
%
% BibTeX users should specify bibliography style 'splncs04'.
%
\bibliographystyle{splncs04}
\bibliography{main}

% ---- Supplementary ----
\clearpage
\appendix
\setcounter{table}{0}
\setcounter{figure}{0}
\renewcommand{\thetable}{A\arabic{table}}
\renewcommand{\theHtable}{A\arabic{table}}% Hyperref figure hyperlink hook
\renewcommand{\thefigure}{A\arabic{figure}}
\renewcommand{\theHfigure}{A\arabic{figure}}% Hyperref figure hyperlink hook

\section*{Appendix}
\vspace{6pt}
\section{Completeness and Balance Scores of Data in the Wild}
\definecolor{highlight}{rgb}{0.92,0.92,0.92}

\begin{table}[th]
\vspace{-21pt}
\caption{We parse three large-scale image-caption datasets, namely VGR, Nocaps, and Flickr30K. Then we compute \textbf{Completeness} and \textbf{Balance} on each corresponding subset involving spatial relations. LR stands for left-right, TB stands for top-bottom, and FB stands for front-behind. ``-complete" means a subsample that is both linguistically and visually complete according to our definitions.}

\centering

\begin{tabular}{@{}lcrrrrrrrr@{}}
\toprule

 &  & \multicolumn{4}{c}{\footnotesize \textbf{Basic Statistics}} & \multicolumn{4}{c}{\footnotesize \textbf{Skew Metrics}}\\
 \cmidrule(lr){3-6} \cmidrule(lr){7-10} 
& \textbf{Dataset} & \#pairs & \makecell{\#unique \\ images} & \makecell{\#unique \\ captions} & \makecell{\#unique \\ concepts} & \textbf{CPL}$_V$ & \textbf{CPL}$_L$  & \textbf{BLC}$_V$ & \textbf{BLC}$_L$  \\ 
 
\toprule

\multirow{5}{*}{\rotatebox[origin=c]{90}{\textbf{VGR}}} & Original & 21,944 & --- & --- & --- & --- & --- & --- & --- \\
& Spatial & 19,419 & 4,348 & 12,163 & 885 & --- &  --- & --- & --- \\
& LR & 15,482 & 3,244 & 9,816 & 695 & 80 & 100 & 93 & 99 \\
& TB & 2,505 & 1,553 & 1,513 & 439 & 66 & 73 & 49 & 62 \\
& FB & 11,162 & 527 & 834 & 209 & 69 & 95 & 58 & 98 \\ 

\rowcolor{highlight} & LR-complete & 14,352 & 3,074 & 8,766 & 414 & 100 & 100 & 96 & 99 \\
\rowcolor{highlight} & TB-complete & 784 & 488 & 395 & 77 & 100 & 100 & 57 & 69 \\
\rowcolor{highlight} & FB-complete & 741 & 335 & 449 & 61 & 100 & 100 & 68  & 100 \\

\midrule

\multirow{4}{*}{\rotatebox[origin=c]{90}{\textbf{Nocaps}}} & Original & 45,000 & --- & --- & --- & --- & --- & --- & --- \\
& Spatial  & 14,583 & 3,827 & 14,241 & 2,975 & --- & --- & --- & --- \\
& TB & 12,726 & 3,540 & 12,714 & 2,273 & 63 & 62 & 35 & 31 \\
& FB & 1,854 & 1,051 & 1,854 & 703 & 65 & 64 & 57 & 45 \\ 

\rowcolor{highlight} & TB-complete & 4,922 & 2,176 & 4,921 & 369 & 100 & 100 & 46 & 48 \\
\rowcolor{highlight} & FB-complete & 718 & 491 & 718 & 110 & 100 & 100 & 75 & 62 \\ 

\midrule

\multirow{4}{*}{\rotatebox[origin=c]{90}{\textbf{Flickr30k}}} & Original & 155,070 & --- & --- & --- & --- & --- & --- & --- \\
& Spatial & 44,918 & 20,784 & 43,419 & 4,205 & --- & --- & --- & --- \\
& TB & 37,541 & 18,436 & 37,462 & 3,003 & 67 & 66 & 36 & 29 \\
& FB & 7,351 & 5,333 & 7,350 & 1,483 & 65 & 63 & 57 & 38 \\

\rowcolor{highlight} & TB-complete & 30,162 & 16,132 & 30,102 & 872 & 100 & 100 & 38 & 31 \\
\rowcolor{highlight} & FB-complete & 4,016 & 3,185 & 4,016 & 265 & 100 & 100 & 66 & 48 \\

\bottomrule

\end{tabular}
\vspace{-6pt}
\label{tab:metrics_data_in_the_wild}
\end{table}

On a high level, we seek to identify the causes for the dissatisfying relation learning in text-to-image generation from the data distribution angle. Our theory shows promising predictivity both on synthetic images and on real images captured in a clean and controlled setting. Ultimately, we expect to extend our investigation to real images \textit{in the wild}. To facilitate research along this vein, we have parsed and analyzed three large-scale image-caption datasets, namely VGR (Visual Genome Relation) \cite{vlm_bow}, Nocaps (Novel Object Captioning) \cite{nocaps}, and Flickr30K \cite{flickr30k}. We draw subsamples by matching spatial phrases in the captions for three spatial relations: left-right (LR), top-bottom (TB) and front-behind (FB). We present \textbf{Completeness} and \textbf{Balance} scores, along with basic statistics for each of these subsamples in Table~\ref{tab:metrics_data_in_the_wild}.

VGR is carefully curated from Visual Genome \cite{VG} for the study of relational understanding in vision-language models. Thus, most of the examples in VGR involve a spatial relation. On the other hand, only a small subset in Nocaps or Flick30k contains spatial information. The supportive data size further shrinks if we constrain our subsamples to be complete under our definition\footnote{We find complete subsamples by iteratively removing concepts with an incomplete support, until all concepts in the remaining data have a complete support.}. As can be seen from highlighted rows in Table~\ref{tab:metrics_data_in_the_wild}, \textbf{the largest complete subsample only constitutes a tiny fraction of the original data}, plus being highly unbalanced. This emphasizes the importance of designing architectures or algorithms to enable generalization even from skewed data sources.

\section{Experiments on Synthetic Data --- Additional Results}

\begin{figure}[h]
\centering
  \includegraphics[width=0.75\textwidth]{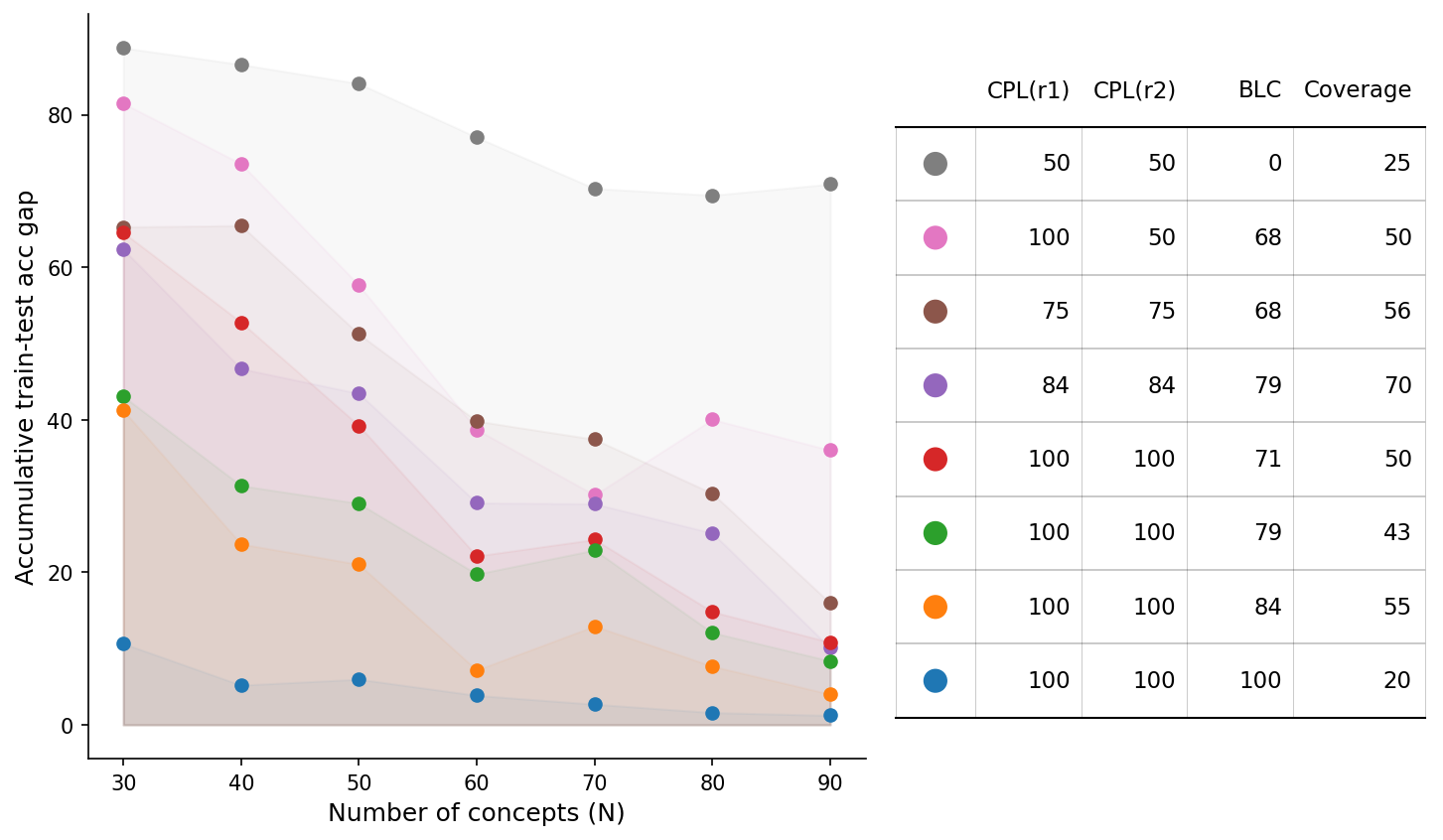}%
  \caption{Under the definition of roles ($r_1$, $r_2$) from \textbf{visual} perspective (``top", ``bottom"), the plot shows final testing accuracy against distributional properties of the training set. Legend and corresponding metrics are summarized in the table. This plot suggests that visual incompleteness and imbalance lead to larger gaps because both the final generalization performance and generalization speed are hampered.}
  \label{figure:image_skew_gap_and_table}
\end{figure}

\begin{figure}[h]
\centering
  \includegraphics[width=12cm]{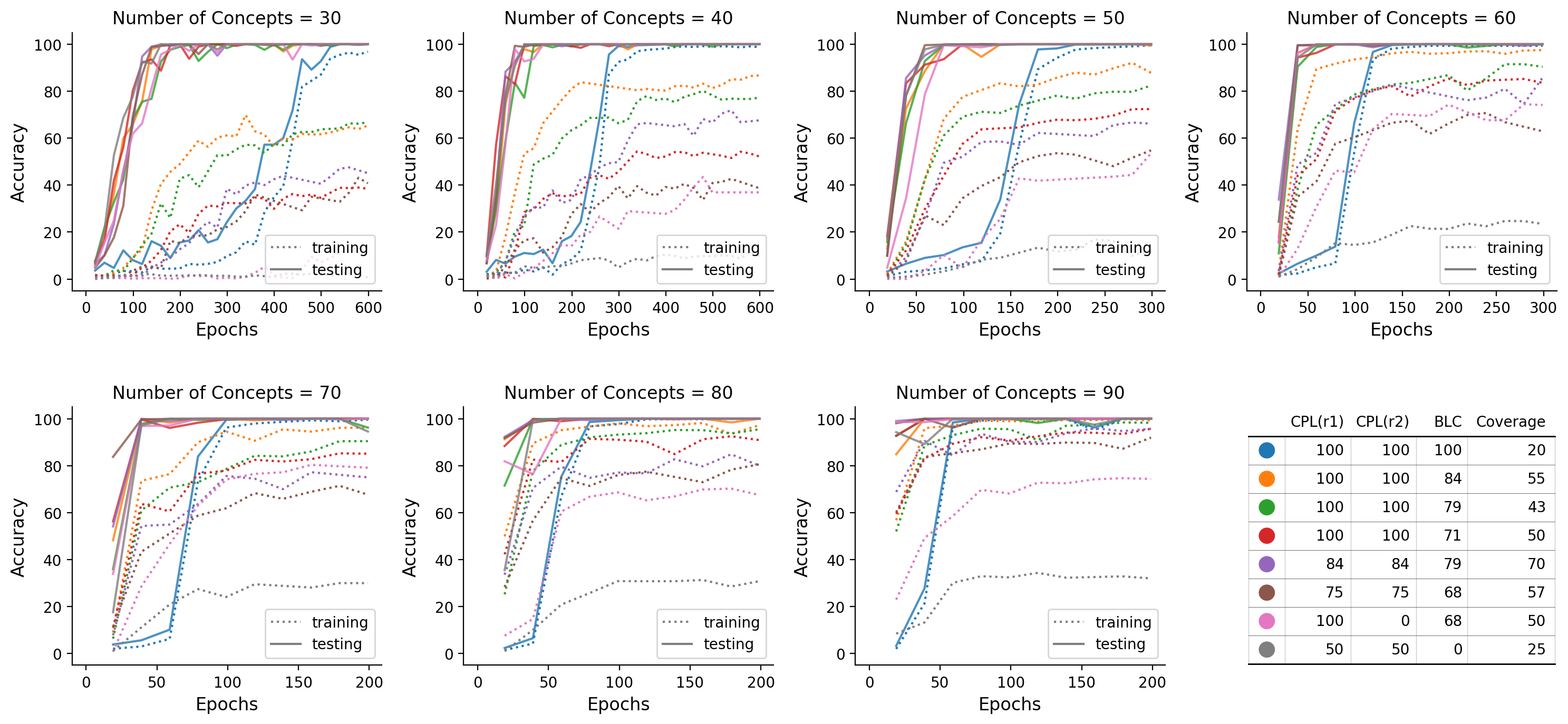}%
  \caption{Detailed results on the impacts of \textbf{Completeness}$_V$ and \textbf{Balance}$_V$. Panels are separated according to the number of concepts. Training accuracy always converges to perfect, whereas testing accuracy is hampered to varying degrees by visual incompleteness or imbalance. This degradation is more severe for a small number of concepts.}
  \label{figure:image_skew_separated}
\end{figure}

\begin{figure}[h]
\centering
  \includegraphics[width=12cm]{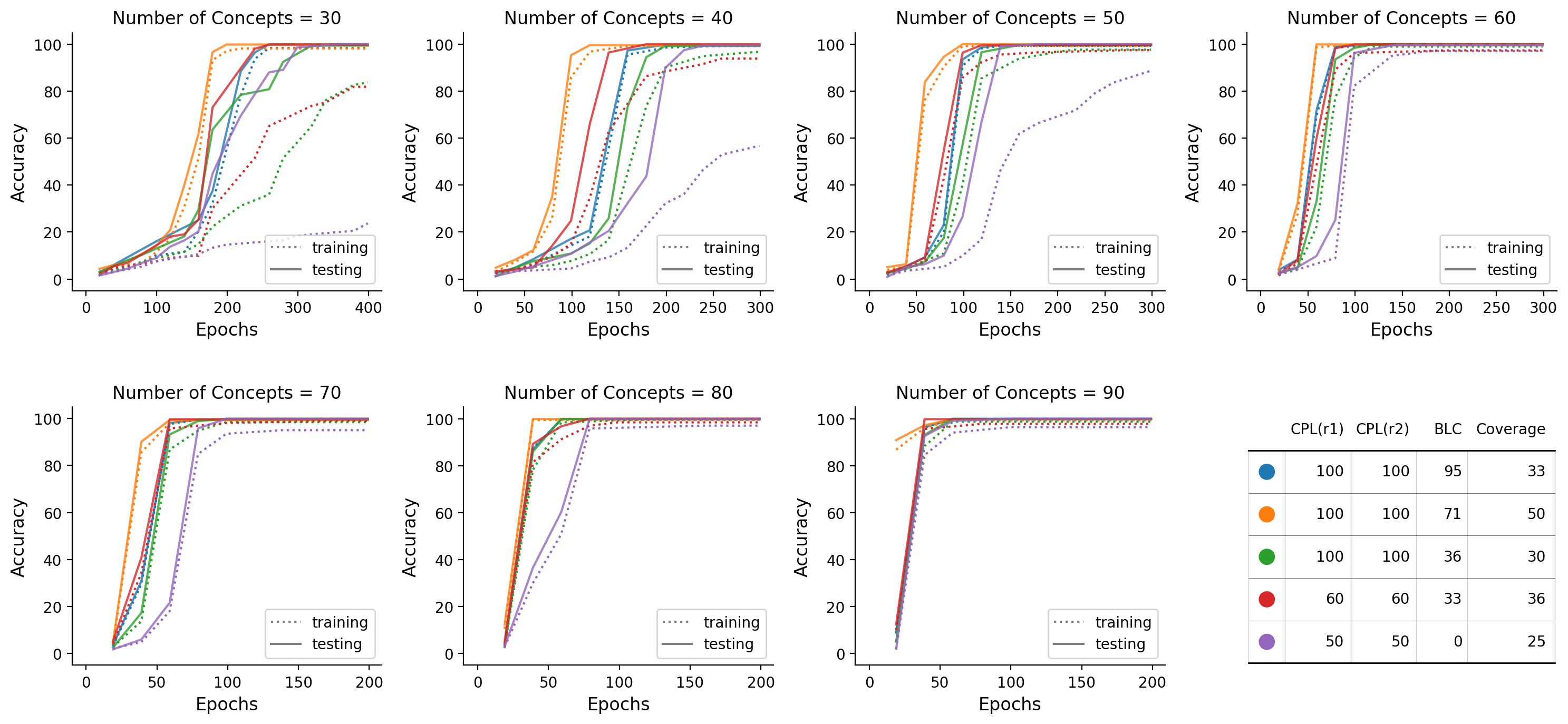}%
  \caption{Detailed results on the impacts of \textbf{Completeness}$_L$ and \textbf{Balance}$_L$. Panels are separated according to the number of concepts. Training accuracy always converges to perfect, whereas testing accuracy is hampered to varying degrees by linguistic incompleteness or imbalance. This degradation is more severe for a small number of concepts.}
  \label{figure:linguistic_skew_separated}
\end{figure}

Figure~\ref{figure:image_skew_gap_and_table} plots for the train-test accuracy gap under varying \textbf{Completeness}$_V$ and \textbf{Balance}$_V$ scores. Moreover, detailed learning curves are comprehensively presented in Figure~\ref{figure:image_skew_separated} and \ref{figure:linguistic_skew_separated}. 

\clearpage

\clearpage
\section{Experiments on Synthetic Data --- Details}
\label{sec:implementation_details_synthetic_images}
\subsection{Icons and Their Names}

\begin{figure}[h]
\centering
  \vspace{-9pt}
  \includegraphics[width=11cm]{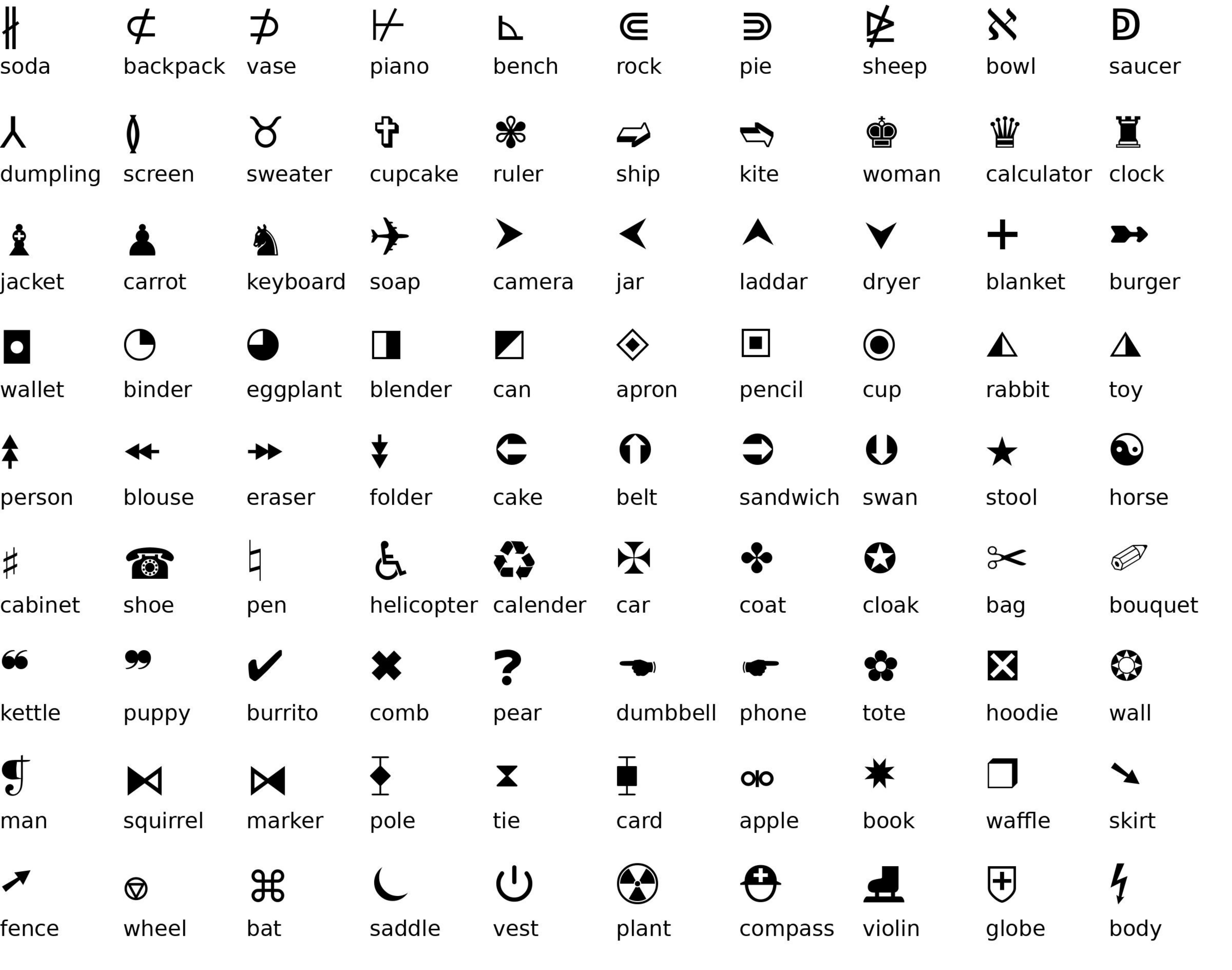}%
  \caption{90 icons paired with 90 common nouns as their names.}
  \vspace{-9pt}
  \label{figure:unicode}
\end{figure}

We consider unicode icons as concepts, each paired with a concrete noun as the identity. To create an image, we draw icons\footnote{Each icon is associated with a font being one of \textit{DejaVuSans}, \textit{DejaVuSansMono}, and \textit{Symbola}. We use three fonts because no single font covers all 90 concepts.} on a white canvas. Figure~\ref{figure:unicode} shows 90 individual icons which we leverage in our synthetic experiments. 

\subsection{Model Configs}
\vspace{-9pt}

\small{
\begin{lstlisting}

    layers_per_block = 2
    block_out_channels = (64, 256, 1024)
    down_block_types = (
        "DownBlock2D",
        "CrossAttnDownBlock2D",
        "CrossAttnDownBlock2D",
    )
    up_block_types = (
        "CrossAttnUpBlock2D", 
        "CrossAttnUpBlock2D",
        "UpBlock2D",
    )
    cross_attention_dim = 128
    patch_size = 2
    conv_in_kernel = 3
    conv_out_kernel = 3
\end{lstlisting}
}

\vspace{-6pt}

\subsection{Training Configs}

We first pretrain the model on single-icon crops to familarize it with the appearance of each icon. Then we finetune the model to generate two icons with different training set statistics.

\vspace{3pt}

Pretraining: 
\vspace{-6pt}
\small{
\begin{lstlisting}
    image_size = (32,32)
    mixed_precision = "fp16" 
    noise_schedule = "squaredcos_cap_v2"
    learning_rate = 1e-4
    lr_warmup_steps = 1000
    train_batch_size = 16
    num_train_timesteps = 100 # diffusion steps
\end{lstlisting}
}

Finetuning:
\vspace{-6pt}
\small{
\begin{lstlisting}
    image_size = (64,32) # H, W
    mixed_precision = "fp16" 
    noise_schedule = "squaredcos_cap_v2"
    learning_rate = 1e-4
    lr_warmup_steps = 1000
    train_batch_size = 16
    num_train_timesteps = 100 # diffusion steps
\end{lstlisting}
}

\vspace{-6pt}

\subsection{Hyperparameter Search}

layers\_per\_block $\in$ \{2, 3, 4\}.\\
block\_out\_channels $\in$ \{(32, 128, 512), (128, 256, 512), (64, 256, 1024), (64, 512, 2048)\}.\\
learning\_rate $\in$ \{1e-3, 5e-4, 1e-4, 1e-5\}.\\
train\_batch\_size $\in$ \{8, 16, 64\}.\\
lr\_warmup\_steps $\in$ \{100, 1000, 2000, 5000\}.\\

\vspace{-6pt}

\section{Experiments on Natural Data --- Details}
\label{sec:implementation_details_natural_images}

\subsection{Model Configs}

\small{
\begin{lstlisting}
    layers_per_block = 2
    block_out_channels = (512, 512, 1024) 
    down_block_types = (
        "DownBlock2D",
        "CrossAttnDownBlock2D",
        "CrossAttnDownBlock2D",
    )
    up_block_types = (
        "CrossAttnUpBlock2D", 
        "CrossAttnUpBlock2D",
        "UpBlock2D",
    )
    cross_attention_dim = 128
    patch_size = 2
    conv_in_kernel = 3
    conv_out_kernel = 3
\end{lstlisting}
}

\subsection{Training Configs}

Again there are two training stages. At the first stage the model learns to synthesize single-object crops. At the second stage the model learns to synthesize two objects in a specified spatial relation. After pretraining, we find it more efficient to only finetune weights in the projection and attention layers, without loss of performance compared to finetuning all parameters. Also, a larger finetuning learning rate leads to rapid convergence without hurting performance.

\vspace{3pt}

Pretraining: 
\vspace{-6pt}
\small{
\begin{lstlisting}
    image_size = (32,32)
    mixed_precision = "fp16"  
    noise_schedule = "squaredcos_cap_v2"
    learning_rate = 1e-4
    lr_warmup_steps = 1000
    train_batch_size = 16
    num_train_timesteps = 100 # diffusion steps
\end{lstlisting}
}

Finetuning:
\vspace{-6pt}
\small{
\begin{lstlisting}
    image_size = (32,64) # H, W
    mixed_precision = "fp16"  
    noise_schedule = "squaredcos_cap_v2"
    learning_rate = 5e-4
    lr_warmup_steps = 1000
    train_batch_size = 16
    num_train_timesteps = 100 # diffusion steps
    trainable_parameters = ["attentions", "encoder_hid_proj"]
\end{lstlisting}
}

\vspace{-6pt}
\subsection{Hyperparameter Search}

layers\_per\_block $\in$ \{2, 3, 4\}.\\
block\_out\_channels $\in$ \{(64, 256, 1024), (128, 512, 1024), (512, 512, 1024), (512, 1024, 1536)\}.\\
learning\_rate $\in$ \{1e-3, 5e-4, 1e-4\}.\\
train\_batch\_size $\in$ \{16, 64\}.\\
lr\_warmup\_steps $\in$ \{1000, 2000, 5000, 10000\}.\\
num\_train\_timesteps $\in$ \{100, 500, 1000\}.\\

\section{Probing Experiments on Text Encodings}
\label{sec:probing}

In the main paper, a conceptual framework for text-to-image generation is introduced, containing three components: text encoder, communication channel and image decoder. We have argued that the text-to-image generation performance crucially depends on all three components functioning and coordinating appropriately. Concretely, the text encoder is responsible for representing features extracted from raw text. The communication channel aims at transmitting features cross-modality. Lastly, the image decoder instantiates an output into the pixel space. Note, though we conceptually divide up the pipeline into three components, \textbf{there is no hard boundary for the points at which different functions actually take place}, because all representations and transformations in this pipeline are homogeneously encoded as differentiable parameters. For example, a weak text encoder could under-process the text features, thus expecting the communication channel to further process text features on the fly. While it is convenient to consider three types of responsibilities for developing hypotheses regarding the sources of error, under the hood, these three components in an end-to-end system could easily transfer or take over responsibilities to or from each other.

This raises the question: \textbf{How rich text features from pre-trained text encoders are?} As argued above, richer text features would give the subsequent communication channel an advantage point, lowering the complexity of transformations needed to transmit the key message cross-modality. This intuition aligns with the known benefit of representation learning, which saves efforts for downstream modeling. On the other hand, a less capable text encoder would increase the difficulty of learning the communication channel, and even create obstacles if critical information gets lost. 

\begin{table}[h]
%\parbox{.58\textwidth}{
\centering

  \caption{\footnotesize Probing results, showing that CLIP text encodings do not effectively facilitate subsequent modules to figure out the spatial positions of physical entities. In contrast, T5 and VLC are better candidates for coupling with text-to-image models.}
\centering	
\footnotesize
 \begin{tabular}{@{}lcccc@{}}
\toprule
\multicolumn{2}{c}{\footnotesize Configuration} & \multicolumn{3}{c}{\footnotesize Performance}
\\
 \cmidrule(lr){1-2} \cmidrule(lr){3-5}   
{ \tiny Text Encoder}  &  \tiny \makecell{MLP \\ Num\_layers} & \tiny Tr Acc  &   \tiny Val Acc  & \tiny \makecell{Out-distr. \\ Te Acc}  \\
\midrule
     CLIP-B/32 & 1 & 74.8 & 74.8 & 74.4 \\
     CLIP-B/32 & 2 & 74.6 & 74.4 & 74.2 \\
     CLIP-B/32 & 3 & 74.9 & 73.9 & 74.3 \\
     CLIP-L/14 & 1 & 74.6 & 73.5 & 74.3  \\
     CLIP-L/14 & 2 & 74.8 & 74.3 & 74.6 \\
     CLIP-L/14 & 3 & 75 & 74.9 & 74.5 \\
\midrule
     T5-small & 1 & 98.5 & 98.5 & 93.1 \\
     T5-small & 2 & 100 & 100 & 95.8 \\
     T5-small & 3 & 100 & 100 & 96.7 \\
     T5-flan-xxl & 1 & 99.9 & 99.8 & 99.3 \\
     T5-flan-xxl & 2 & 100 & 100 & 99.8 \\
     T5-flan-xxl & 3 & 100 & 100 & 99.4 \\
\midrule
     VLC-L/16 & 1 & 99.1 & 99.1 & 96.1 \\
     VLC-L/16 & 2 & 99.9 & 100 & 96.0 \\
     VLC-L/16 & 3 & 100 & 100 & 99.8 \\
     
\bottomrule
\end{tabular}
  \vspace{-9pt}
  \label{tab:probing}
%}
\end{table}

To this end, we perform probing experiments to analyze how good an advantage point a text encoder can provide for learning a communication channel. We consider three pre-trained text encoders: CLIP \cite{clip}, T5 \cite{T5} and VLC \cite{vlc}, which are chosen to be representative of different model families and pre-training objectives. CLIP is an encoder-only dual architecture pre-trained with image-text contrastive loss. T5 is an encoder-decoder architecture pre-trained with the causal language modeling loss. VLC is an encoder-only multimodal architecture pre-trained with masked language modeling, masked image modeling as well as image-text matching objectives. Pertaining to our work, the core responsibility of the communication channel is to translate linguistic fillers and roles to the corresponding visual entities and spatial positions. Thus, we train probing classifiers to label the noun tokens in a caption with spatial positions (e.g. top or bottom). For example, given a caption ``a book is on top of a cup”, we apply an MLP on the encoded features of the ``book” and ``cup” tokens, and expect the MLP to output label 0 for ``book” and label 1 for ``cup”. For another example, “a laptop is at the bottom of a phone”, the MLP should output label 1 for ``laptop” and label 0 for ``phone”. \textbf{The probing performance signifies the utility of a pre-trained text encoder in assisting a diffusion model, particularly in the context of generating spatial relations}. Though a probing classifier is a drastic simplification, we believe that it still captures the main problem.

We select 200 common English nouns meaning concrete objects, then divide them into 170 “in-distribution” and 30 “out-of-distribution” nouns. We expect a successful probe to generalize to out-of-distribution nouns, which would strongly imply the existence of robust feature directions encoding spatial information. We train 1-, 2-, and 3-layer MLPs with ReLU nonlinearities. The 1-layer MLP is equivalent to linear probing. The MLPs are trained for 10-20 epochs until convergence. We observe consistent results across random data-splitting outcomes (5 seeds) and hyperparameters such as learning rate (\{1e-3, 1e-4\}) and batch size (\{16, 64\}). Table \ref{tab:probing} summarizes our results. CLIP struggles to exceed 75\% and this cannot be improved by increasing the encoder size or the number of MLP layers. T5-small achieves perfect in-distribution performance, but makes minor mistakes when generalizing out-of-distribution. T5-flan-xxl \cite{flan-t5} achieves perfect accuracy in all three splits. VLC performs on par with T5-flan-xxl in-distribution, and slightly underperforms T5-flan-xxl out-of-distribution.
 
\begin{figure}[h]
\centering
  \includegraphics[width=12cm]{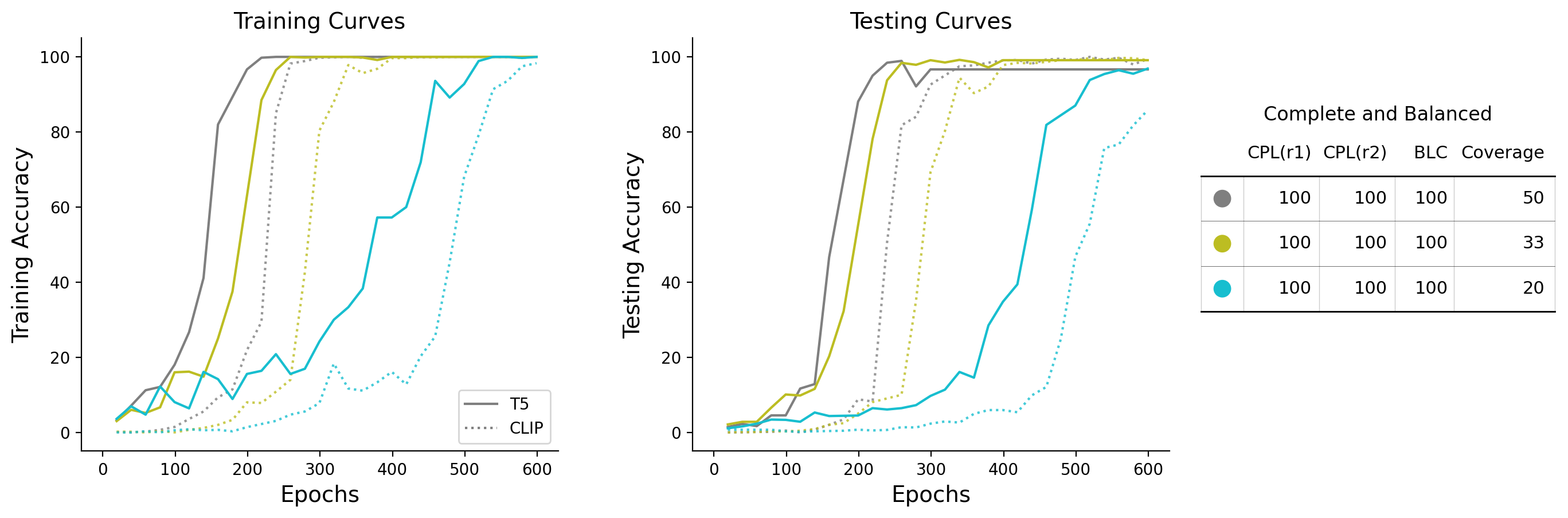}%
  \caption{Ablation study on the text encoder while keeping all other training configurations fixed. The performance of diffusion paired with CLIP text encoder consistently falls behind the performance with T5, suggesting that CLIP would indeed make learning harder and slower.}
  \label{figure:ablation_clip}
\end{figure}

Our probing experiments undermine CLIP-text-encoder’s effectiveness in facilitating subsequent modules to figure out the spatial positions of physical entities. In contrast, T5 and VLC are better candidates for coupling with text-to-image models. Indeed, our ablation study on diffusion models confirms this claim. As shown in Figure~\ref{figure:ablation_clip}, all experiments follow the same setting except that solid lines represent the T5-small text encoder and dotted lines represent the CLIP-B/16 text encoder. It is evident that dotted lines consistently fall behind solid lines. Note, \textbf{our analysis does not render CLIP as incapable of serving as the text encoder for text-to-image models. Nevertheless, compared to T5 and VLC, CLIP might have to offload more burden to the communication channel for nontrivial processing.} 

%\clearpage
\vspace{-5pt}

%\vspace{-12pt}
\section{Ablation Results on Image Positional Embeddings}
\label{sec:img_posemb_ablation}

In this section, we investigate a crucial factor --- image positional embeddings --- that draws a connection from spatial features to the correct absolute positions in the 2D image coordinates. Without explicitly embedding patch positions, such ability could be largely damaged. Though, arguably, zero-paddings in the convolutional layers could provide complementary positional information, we believe that this cannot fully compensate for the loss incurred by a lack of explicit image positions. \textbf{We perform ablation studies where the image positional embeddings are removed while remaining other architectural components intact}. Specifically, we set the number of concepts to be 30 because a smaller number of concepts imposes a greater challenge such that discrepancies in performance can be easily revealed. 

\begin{figure}[!t]
\centering
  \includegraphics[width=\textwidth]{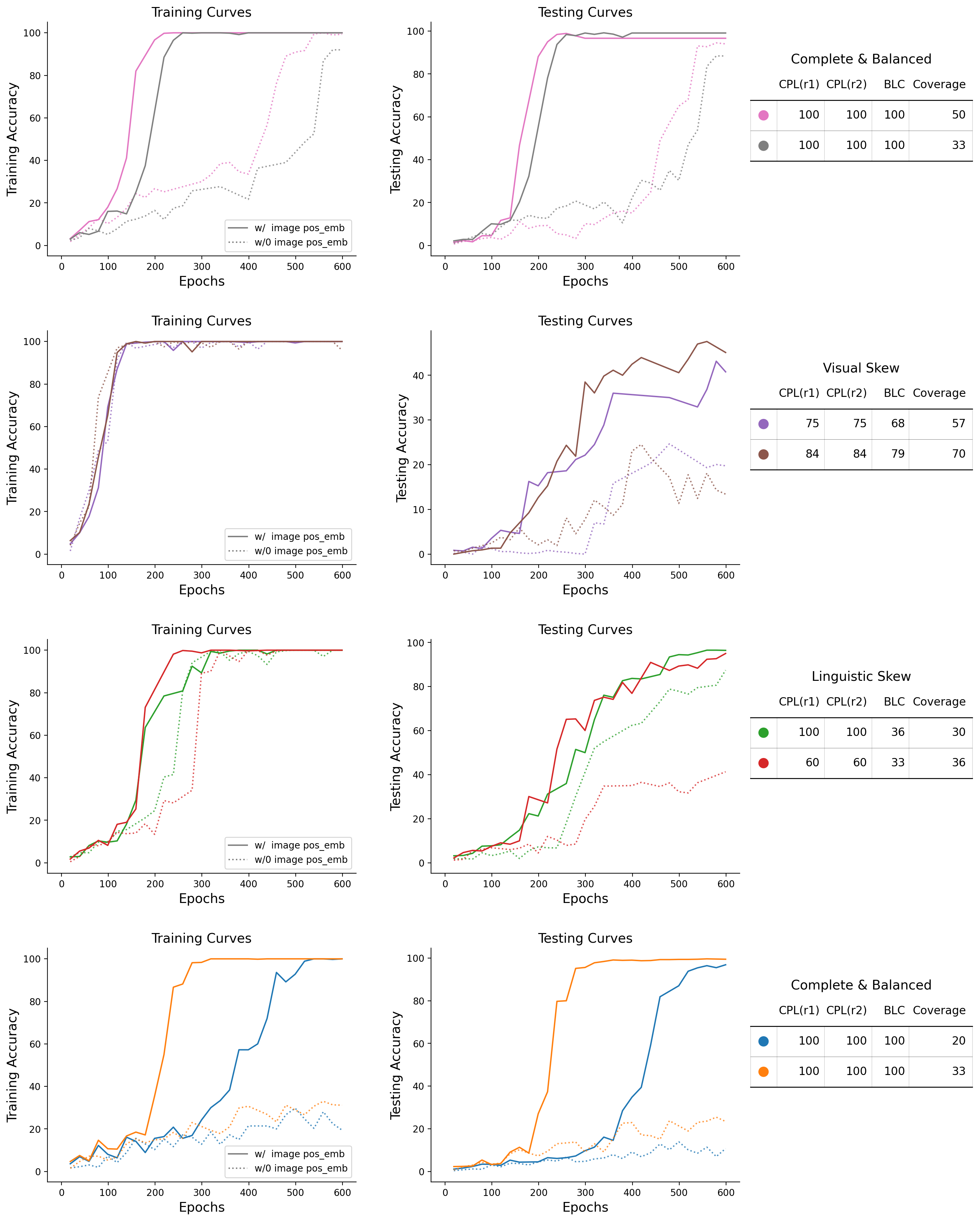}%
  \caption{Ablation study on whether to include image positional embeddings. We see that diffusion models without the image positional embedding are inferior to their counterparts across all panels, albeit failing or underperforming for different reasons.}
  \vspace{-15pt}
  \label{figure:ablation_image_pos_emb}
\end{figure}

Figure~\ref{figure:ablation_image_pos_emb} presents the results of this ablation study. For notational convenience, we call models \textit{with} image positional embeddings the ``Pos-models", and call models \textit{without} image positional embeddings the ``NoPos-models". It can be observed that, NoPos-models perform worse than their counterpart Pos-models across all panels. Comparing the panels between rows, we identify different types of failure. In the first row (pink and grey), NoPos-models exhibit much slower convergence speed, but are still able to converge and generalize eventually. In the second row (purple and brown), NoPos-models do not fall behind in terms of training accuracy. Yet they are generalizing to the testing set much worse than Pos-models. In the third row (green and red), there are small gaps between Pos-models and NoPos-models in training accuracy. And this gap widens regarding testing accuracy. In the fourth row (blue and orange), unfortunately, NoPos-models have trouble even fitting the training set.

We attempt to associate the failure patterns of NoPos-models with our proposed linguistic or visual skew in the training data. It appears that in rows 2\&3, models are hampered by the skew, with NoPos-models being more severely hampered. When the training data is complete and balanced, the NoPos-models would generalize on larger coverages (row 1), albeit at much slower paces, and fail spectacularly on smaller coverages (row 4). We suspect that the interplay between the data skew and the failure pattern of NoPos-models is intricate, so we have not reached a conclusive statement. But the key takeaway is clear: \textbf{removing image positional embeddings leads to notable performance degradation in all cases}. 

\clearpage
\section{Latent Diffusion Experiments}
\label{sec:latent_experiments}

\begin{figure}[h]
\centering
  \vspace{-24pt}
  \includegraphics[width=\linewidth]{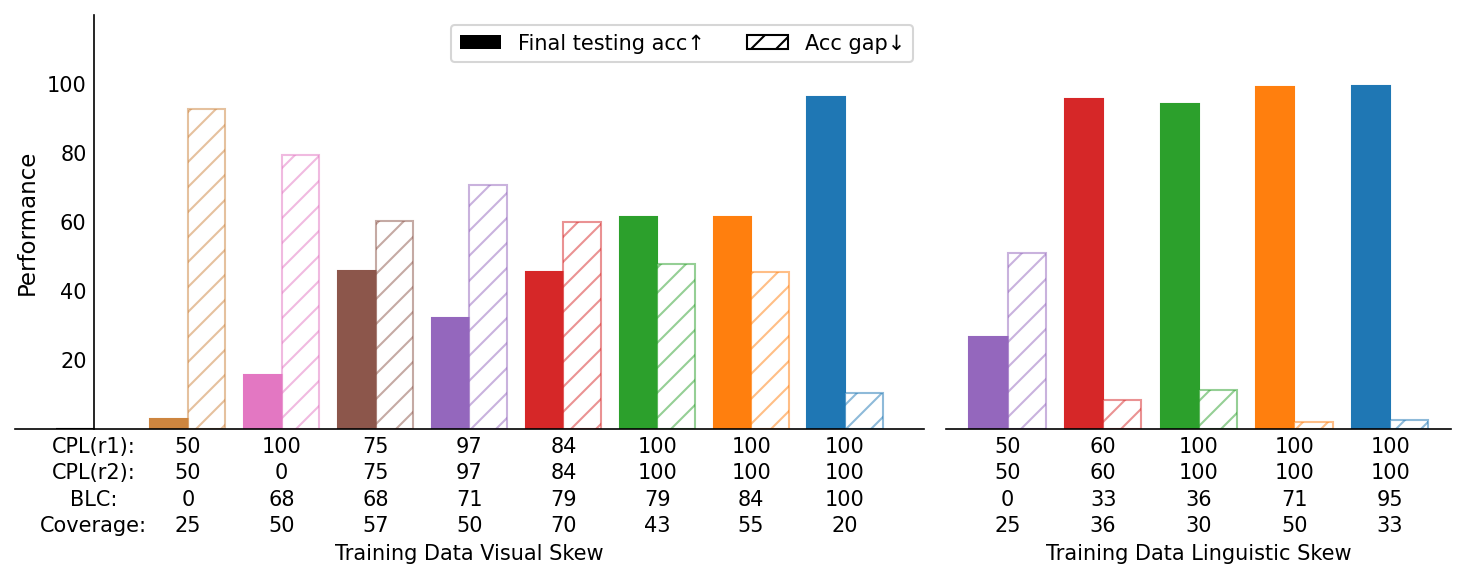}\\
  \vspace{10pt} 
  \includegraphics[width=\linewidth]{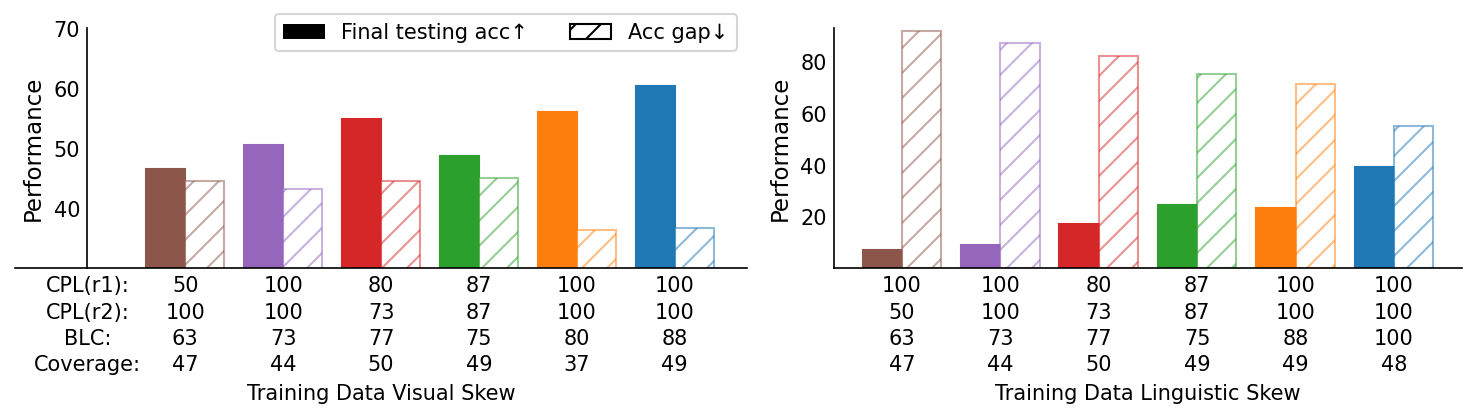}
  \caption{Results for latent diffusion models. A pretrained VAE is frozen, while the UNet is trained from scratch. On both synthetic (top) and natural (bottom) images, we observe trends consistent with our pixel diffusion findings --- lower CPL and BLC harms testing accuracy and widens the generalization gap, while Coverage is less predictive.}
  \vspace{-12pt}
  \label{fig:latent_diffusion_results}
\end{figure}

We repeat the experiments on synthetic and natural images with a latent diffusion model, in order to validate that our arguments do not break when moving from the pixel to a latent space. We adopt a pretrained VAE (stabilityai/stable-diffusion-2-1), freeze it, and train the UNet from scratch. We use the same set of hyperparameters as in Appendix~\ref{sec:implementation_details_synthetic_images} and \ref{sec:implementation_details_natural_images}, except for two modifications: 1) input resolution is increased by a factor of four, 2) we use four channels to match the input/output dimensions of the VAE, as opposed to three for rgb inputs. Results are plot in Figure~\ref{fig:latent_diffusion_results}, which are highly consistent with our pixel diffusion findings.

\vspace{-6pt}
\section{Auto-Evaluation of Relations}
\label{sec:autoeval}

Our experiments critically depend on an auto-eval engine that correctly judges both the object identity and their spatial relations. However, current evaluation methods of generated images have lingering issues of not paying attention to relations. Off-the-shelf open-vocabulary detection models have problems such as insensitivity to certain object classes, redundant predictions and poor reliability at low resolutions. Thus, we decide that they are not suitable for our purpose. We then try to leverage vision-language models, inspired by \cite{T2I-compbench, llmscore}, which ended up not working. Eventually, we arrive at a working solution via finetuning ViT \cite{ViT}, despite making certain compromises, i.e. assuming that objects only occur at fixed image regions. \textbf{This section describes our attempts at prompting VLMs and finetuning ViT for automatic evaluation, pertaining to experiments on natural images}.

\subsection{Leveraging Large Vision-Language Models}

\textbf{Zero-Shot Multiple-Choice VQA} Our first attempt was using off-the-shelf VLMs --- BLIP2 \cite{li2023blip} and LLaVA1.5 \cite{liu2023improved} --- to evaluate spatial relations, since they have been trained on position-related data (e.g. Visual Genome \cite{VG}). We phrase the prompt as a zero-shot multiple-choice question. For example, 

\small{
\begin{lstlisting}
    Which caption better describes the image? 
    A. A mug to the right of a knife;
    B. A mug infront of a knife;
    C. A mug behind a knife; 
    D. A mug to the left of a knife.  
\end{lstlisting}
}

Table \ref{tab:vqa} shows that both LLaVA1.5-VQA and BLIP2-VQA barely outperformed 25\% on the 4-way classification task. %Hence, we opted not to pursue further improvements on multiple-choice VQA tasks.

\begin{table}[h]
\vspace{-9pt}
\parbox{.48\textwidth}{
\caption{Zero-Shot Multiple-Choice VQA Results of BLIP2 and LLaVA1.5 on the What's Up Benchmark.}
\centering
\begin{tabular}{cc}
\toprule
Model & Accuracy \\
\midrule
BLIP2 & 0.26 \\
LLaVA1.5 & \textbf{0.28} \\
\bottomrule
\end{tabular}
%\vspace{5pt}
\label{tab:vqa}
}
\hfill
\parbox{.48\textwidth}{
\caption{Zero-Shot Object Detection Results of BLIP2 and LLaVA1.5 on the What's Up Benchmark.}
\centering
\begin{tabular}{ccc}
\toprule
 & BLIP2 & LLaVA1.5\\
\midrule
Acc. on True Labels & \textbf{100\%} &  83.6\% \\
Acc. on False Labels & 12.1\% & \textbf{94.2\%} \\
Average Accuracy &  16.9\% & \textbf{93.6\%} \\
\bottomrule
\end{tabular}
%\vspace{5pt}
\vspace{-9pt}
\label{tab:object}
}
\end{table}

\textbf{Zero-shot Object Detection} Though VLMs struggle to recognize positional relationships, they do excel in detecting the existence of objects. VLMs could still be helpful if the generated images were preprocessed into single-object crops, under the assumption that objects only occur in fixed image regions and cropping does not cut through objects. For example, in our setting with binary spatial relations and a 32x64 input size, preprocessing involves cropping the image into two 32x32 squares. We then feed the two crops into the VLM in two individual passes, together with the prompt: ``\texttt{Is a [object] in the image?}'', where we loop through all 18 object classes in the What'sUp benchmark. As shown in Table~\ref{tab:object}, BLIP2 tends to affirm the presence of any queried object, leading to 100\% accuracy for true labels but a significantly worse accuracy, 12.1\%, for false labels, indicating a high false-positive rate. Given that there are 18 objects in total, questions with false labels outnumber those with true labels by 17 times. This explains why the average accuracy is close to that for false labels. 

In contrast, LLaVA1.5 remarkably outperforms BLIP2 with much more balanced false-positive and false-negative rates. Therefore, we proceed to compute the frequency when LLaVA judges correctly for both crops given a generated image.
%Given LLaVA's impressive performance in object detection, we explored using LLaVA for positional detection in a manual manner. The data processing procedure is the same as that in object detection.  Subsequently, we ask, ``\texttt{Is a [object1] in image1?}'' and ``\texttt{Is a [object2] in image2?}'' A classification is deemed correct if both questions are affirmed.   
We find certain object classes particularly difficult for LLaVA. The overall accuracy is 68.14\%. Excluding difficult classes progressively improves accuracy to 92\% (Table~\ref{table:hard_to_detect_classes}). 
%Removing the bottom two classes (tape and flower) elevates accuracy to about 83\%, and eliminating the bottom three (tape, flower, and knife) boosts it to around 87\%. Further exclusion of the bottom four classes, including fork or scissors, increases accuracy to 89\%, and removing the bottom five classes (tape, flower, knife, fork, and scissors) results in an accuracy of 92\%.
However, we are not satisfied with LLaVA as an auto-eval engine, because we expect near-oracle accuracy. We list the top10 frequent misclassifications in Table \ref{table:item_pairs}. Understandably, LLaVA tends to confuse between semantically similar objects.

\begin{table}[h]
\centering
%\vspace{-9pt}
\caption{LLaVA1.5 Zero-Shot Object Detection has highly unbalanced accuracy between easy and hard classes}
\small
\begin{tabular}{lc}
\toprule
Inclusion of classes & Accuracy \\
\midrule
All & 0.68 \\
Exclude the worst 2 Classes (Tape, Flower) & 0.83 \\
Exclude the worst 3 Classes (Tape, Flower, Knife) & 0.87 \\
Exclude the worst 4 Classes (Tape, Flower, Knife, Fork/Scissors) & 0.89 \\
Exclude the worst 5 Classes (Tape, Flower, Knife, Fork, Scissors) & 0.92 \\
\bottomrule
\end{tabular}
%\vspace{-15pt}
\label{table:hard_to_detect_classes}
\end{table}

\begin{table}[h]
\centering
\caption{LLaVA1.5 Zero-Shot Object Detection: Top 10 frequent misclassifications}

\begin{tabular}{lr}

\toprule
(Correct, Predicted) & Frequency \\
\midrule
(Mug, Cup) & 68 \\
(Can, Cap) & 36 \\
(Bowl, Plate) & 28 \\
(Sunglasses, Headphones) & 27 \\
(Remote, Headphones) & 27 \\
(Scissors, Remote) & 25 \\
(Scissors, Headphones) & 24 \\
(Scissors, Cap) & 21 \\
(Flower, Candle) & 21 \\
(Candle, Cap) & 19 \\
\bottomrule
\end{tabular}
%\vspace{-15pt}
\label{table:item_pairs}
\end{table}

\subsection{Crop and Classify}
\begin{figure}[h]
\centering
  \vspace{-6pt}
  \includegraphics[width=\textwidth]{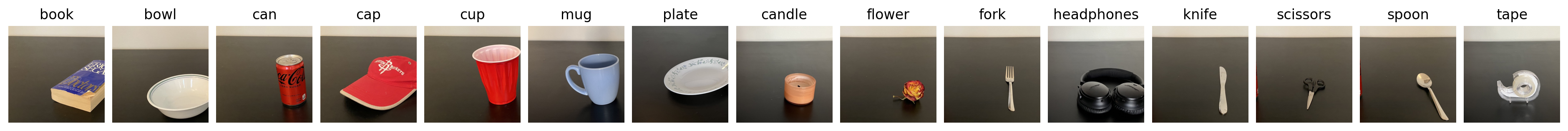}%
  \caption{Finetuning data examples. We finetune ViT to classify objects in image crops for the purpose of automatically evaluating generated images.}
  \vspace{-15pt}
  \label{figure:finetune_vit_examples}
\end{figure}

For the purpose of automatically evaluating the generated images on the What'sUp benchmark, we finally arrive at a working solution that consists of two steps. The first step obtains single object crops from an image based on heuristics such that the dividing line between two objects aligns with the center of the image. The second step classifies the object into one of the 15 classes. We obtain an Oracle classifier via finetuning ViT-B/16\footnote{google/vit-base-patch16-224-in21k}. We use individual object crops from the original images as our finetuning data, splitting into training and validation sets with the ratio 90:10. Examples are illustrated in Figure~\ref{figure:finetune_vit_examples}. We adopt the Adam optimizer with lr=2e-4 and a batch size of 16. The classification accuracy reaches perfect within 5 epochs. For sanity check, we manually examined 144 generated examples and our ViT evaluated correctly on all of them.

\section{Limitations and Future Work}
\label{sec:limitation_and_future}
\normalsize
This work bridges a significant gap by formally characterizing the phenomenological space in text-to-image generation datasets. Though we mainly focus on small-scale and controllable settings for proof-of-concept experiments, a breadth of expansions are enabled as direct next steps. 

\vspace{5pt}
\noindent \textbf{Larger-scale data} The scaling property of our argument remains to be testified in future work. Current difficulties in scaling up lie in the lack of clean data with entity-relation annotations. Controlled experiments require datasets spanning the incomplete $\rightarrow$ complete and imbalanced $\rightarrow$ balanced spectra. One path is to draw subsamples, filtering with relations of interest and for objects bound to different roles. We found Flickr and Nocaps too small after filtering. As seen in Table~\ref{tab:metrics_data_in_the_wild}, the number of images drops to 16k and 2k for Flickr and Nocaps, respectively, for the left-right relation. These will further drop if we desire a high balance score. Our work motivates the development of either cleaner datasets or augmentation techniques to study with perceptually rich images.

\vspace{5pt}
\noindent \textbf{Complex relations} There is ample room for future study to explore beyond binary, grammatical or spatial relations. We see three possible directions: 1) study relations involving $>$2 roles, e.g. ``forming a line/triangle"; 2) distinguish the directionality of a relation, and extend to non-directional relations, e.g. ``next to"; 3) expand to interaction-based relations, usually entailed by verbs, e.g. ``serve", ``feed", ``approach" --- more abstract and dynamic relations may require other computer vision advances such as action/posture recognition, as well as incorporating a temporal axis.

\vspace{5pt}
\noindent \textbf{Guiding data pruning} Establishing links between data distributional properties and models' generalizing behavior not only benefits model understanding and development, but also potentially inspires data pruning techniques to save compute and obtain better performance. In fact, experiments in Section~\ref{sec:experiments_on_natural_images} exhibit such a flavor: we did not use the full What'sUp dataset containing 18 objects, but discarded frequent objects that worsen the skew. Future work may develop pruning strategies informed by phenomenological measurements and evaluate the data-efficiency gain.

% ------------------------------------------------------------------
% Comment everything below if this file is imported by the main file.
% \bibliographystyle{splncs04}
% \bibliography{supp}

% \end{document}

\end{document}